\documentclass{elsart}               

\usepackage{graphicx}          

\usepackage{amsfonts}
\usepackage{amsmath}
\usepackage{booktabs}
\usepackage{placeins}
\usepackage{subcaption}
\usepackage{bm}
\usepackage{siunitx}
\usepackage{fp}
\usepackage{tikz}

\usepackage{color,soul}
\usepackage{xcolor}

\newtheorem{theorem}{Theorem}[section]
\newtheorem{proposition}[theorem]{Proposition}

\newtheorem{assumption}[theorem]{Assumption}

\newcommand{\R}{{\mathbb{R}}}

\newcommand{\Rn}{{\mathbb{R}}^{n}}
\newcommand{\Rm}{{\mathbb{R}}^{m}}

\newcommand{\Rmm}{{\mathbb{R}^{m \times m}}}

\newcommand{\T}{{\text{T}}}     

\newcommand{\eye}{{\boldsymbol{I}}}
\newcommand{\zero}{{\boldsymbol{0}}}

\newcommand{\inv}[1]{{#1}^{-1}}

\newcommand{\RKHS}{\mathcal{H}_{K}}
\newcommand{\innerproduct}[2]{\left\langle #1 , #2 \right\rangle}

\newcommand{\beq}{\begin{equation}}
\newcommand{\eeq}{\end{equation}}
\newcommand{\bb}{\begin{bmatrix}}
\newcommand{\eb}{\end{bmatrix}}

\DeclareMathOperator{\spn}{span}

\newcommand{\hboks}{\hfill$\square$\break}

\newcommand{\boldalpha}{\bm{\alpha}}
\newcommand{\boldbeta}{\bm{\beta}}

\newcommand{\boldepsilon}{\bm{\epsilon}}
\newcommand{\boldzeta}{\bm{\zeta}}
\newcommand{\boldeta}{\bm{\eta}}

\newcommand{\boldsigma}{\bm{\sigma}}
\newcommand{\boldtau}{\bm{\tau}}

\newcommand{\phidot}{\Dot{\phi}}

\newcommand{\phiddot}{\Ddot{\phi}}

\newcommand{\boldPhi}{\bm{\Phi}}
\newcommand{\boldPsi}{\bm{\Psi}}

\newcommand{\bolda}{{\bm{a}}}

\newcommand{\boldc}{{\bm{c}}}

\newcommand{\bolde}{{\bm{e}}}
\newcommand{\boldf}{{\bm{f}}}
\newcommand{\boldg}{{\bm{g}}}
\newcommand{\boldh}{{\bm{h}}}

\newcommand{\boldp}{{\bm{p}}}
\newcommand{\boldq}{{\bm{q}}}
\newcommand{\boldr}{{\bm{r}}}

\newcommand{\boldu}{{\bm{u}}}
\newcommand{\boldv}{{\bm{v}}}
\newcommand{\boldw}{{\bm{w}}}
\newcommand{\boldx}{{\bm{x}}}
\newcommand{\boldy}{{\bm{y}}}
\newcommand{\boldz}{{\bm{z}}}

\newcommand{\xdot}{\Dot{x}}
\newcommand{\ydot}{\Dot{y}}

\newcommand{\xddot}{\Ddot{x}}
\newcommand{\yddot}{\Ddot{y}}

\newcommand{\boldedot}{\Dot{\bm{e}}}

\newcommand{\boldqdot}{\Dot{\bm{q}}}

\newcommand{\boldxdot}{\Dot{\bm{x}}}
\newcommand{\boldydot}{\Dot{\bm{y}}}

\newcommand{\boldqddot}{\Ddot{\bm{q}}}

\newcommand{\boldyddot}{\Ddot{\bm{y}}}

\newcommand{\boldytilde}{\tilde{\bm{y}}}

\newcommand{\boldytildedot}{\Dot{\tilde{\bm{y}}}}

\newcommand{\boldcbar}{\bar{\bm{c}}}

\newcommand{\boldgbar}{\bar{\bm{g}}}

\newcommand{\boldB}{{\bm{B}}}
\newcommand{\boldC}{{\bm{C}}}
\newcommand{\boldD}{{\bm{D}}}
\newcommand{\boldE}{{\bm{E}}}

\newcommand{\boldJ}{{\bm{J}}}
\newcommand{\boldK}{{\bm{K}}}

\newcommand{\boldM}{{\bm{M}}}

\newcommand{\boldP}{{\bm{P}}}

\newcommand{\boldR}{{\bm{R}}}

\newcommand{\boldY}{{\bm{Y}}}

\newcommand{\Qdot}{\Dot{Q}}

\newcommand{\boldDbar}{\bar{\bm{D}}}

\newcommand{\calN}{{\mathcal{N}}}

\newcommand{\ba}{\left[\begin{array}}
\newcommand{\ea}{\end{array}\right]}
\newcommand{\bbox}{\begin{tcolorbox}[colback=gray!10!white,colframe=bcolor]}
\newcommand{\ebox}{\end{tcolorbox}}

\newcommand{\bsa}{\left\{\begin{array}{c}}
\newcommand{\esa}{\end{array}\right\}}

\newcommand{\CC}{{\mathbb C}}
\newcommand{\EE}{{\mathbb E}}

\newcommand{\av}{{\bm a}}

\newcommand{\f}{{\bm f}}
\newcommand{\g}{{\bm g}}
\newcommand{\h}{{\bm h}}

\newcommand{\p}{{\bm p}}

\newcommand{\q}{{\bm q}}

\newcommand{\rv}{{\bm r}}

\newcommand{\vv}{{\bm v}}

\newcommand{\wv}{{\bm w}}

\newcommand{\x}{{\bm x}}
\newcommand{\yv}{{\bm y}}
\newcommand{\z}{{\bm z}}

\newcommand{\A}{{\bm A}}

\newcommand{\B}{{\bm B}}

\newcommand{\Dm}{{\bm D}}

\newcommand{\F}{{\bm F}}

\newcommand{\I}{{\bm I}}

\newcommand{\M}{{\bm M}}

\newcommand{\alphav}{{\bm \alpha}}

\newcommand{\epsv}{{\bm \epsilon}}

\newcommand{\phiv}{{\bm\phi}}

\newcommand{\Phim}{{\bm{\Phi}}}

\newcommand{\yt}{\tilde{\yv}}

\newcommand{\alphavt}{{\tilde{\alphav}}}

\newcommand{\alphavh}{\hat{\alphav}}

\newcommand{\tr}{{\rm T}}

\newcommand{\frach}{{\frac{1}{2}}}

\newcommand{\fw}[1]{}

\begin{document}

\begin{frontmatter}

\title{Nonparametric adaptive payload tracking for an offshore crane 
} 

\author[MTP]{Torbj{\o}rn Smith\corauthref{corresponding}}\ead{torbjorn.smith@ntnu.no}, \author[MTP]{Olav Egeland}\ead{olav.egeland@ntnu.no}

\address[MTP]{Department of Mechanical and Industrial Engineering\\ Norwegian University of Science and Technology (NTNU)\\ 7491 Trondheim, Norway}

\corauth[corresponding]{Corresponding author.}




\begin{abstract}
A nonparametric adaptive controller is proposed for crane control where the payload tracks a desired trajectory with feedback from the payload position. The controller is based on a novel version of partial feedback linearization where the unactuated crane load dynamics are controlled with the position of the actuated crane dynamics instead of the acceleration. This is made possible by taking advantage of the gravity terms in a new Cartesian model that we propose for the load dynamics. This Cartesian model structure makes it possible to implement a nonparametric adaptive controller which cancels disturbances on the crane load by approximating the effects of unknown disturbance forces and structurally unknown dynamics in a reproducing kernel Hilbert space (RKHS). It is shown that the nonparametric adaptive controller leads to uniformly ultimately bounded errors in the presence of unknown forces and unmodeled dynamics. In addition, it is shown that the proposed partial feedback linearization based on the Cartesian model has certain advantages in payload tracking control also in the non-adaptive case.  The performance of the nonparametric adaptive controller is validated in simulation and experiments with good results.
\end{abstract}


\end{frontmatter}


\section{Introduction}\label{sec:introduction}

Cranes play a vital role in construction, manufacturing and logistics by enabling efficient handling of heavy loads. Crane control is a challenging problem due to nonlinearities and underactuation, and payload oscillations can lead to inefficiency and hazardous situations. Cranes operating offshore compound these issues as they are exposed to severe wind and ocean wave disturbances. Control systems may offer significant improvements in crane operations, such as enhanced safety, reduced operational costs, and improved efficiency.

\subsection{Related work}

There is a large body of work on the automatic control of cranes, with comprehensive reviews presented in \cite{Rahman2003} and \cite{Ramli2017}.  An early work on crane control is presented in \cite{Sakawa1985} where a linear optimal controller and a state observer are used to control a rotary crane.

Feedforward techniques such as input shaping have been extensively studied for application in crane control. This is done by convolving the input with a series of impulses to reduce the pendulum motion of the payload \cite{Cutforth2004}. Input shaping for a tower crane was used in \cite{Blackburn2010}, where the nonlinearities of the system were considered. Flatness-based tracking was used in \cite{Knierim2010} to control an overhead crane. The controller combined feedforward control and state feedback to reduce payload oscillations and to improve tracking performance. Flatness-based control was also used in \cite{Kolar2017} to generate and track minimum time trajectories for a gantry crane.

Model predictive control (MPC) has also been applied in crane control. In \cite{Kimiaghalam2001} a hybrid approach was proposed where feedforward control was used with a nonlinear MPC controller to damp payload oscillations on a shipboard crane subject to wave motion. An MPC controller was used in \cite{Arnold2005} to control a mobile boom crane. The coupled nonlinear dynamic model was linearized along the reference trajectory of the system, approximating the nonlinear optimal control problem using a quadratic programming problem. This allowed for a real-time implementation. This work was extended in \cite{Neupert2010} for a mobile boom crane to achieve tracking and anti-sway control. The controllers were derived using input/output linearization, and smooth trajectories for the controllers were generated from operator commands using an MPC controller. In \cite{Vukov2012}, nonlinear MPC was used to control an overhead crane, performing point-to-point trajectories while varying the cable length of the crane and canceling disturbances. An MPC controller was proposed in \cite{Wu2015} to control a two-dimensional overhead crane while minimizing energy consumption and payload swing angle. In \cite{Smoczek2017}, MPC with a particle swarm optimizer was proposed to control an overhead two-dimensional crane, performing tracking control and parameter identification online while limiting the oscillations of the crane payload. A nonlinear MPC controller was combined with a Lyapunov-based damping controller in \cite{Tysse2022} for tracking control of a knuckle boom crane. The exponentially stabilizing damping controller ensured that the payload oscillation was bounded when the MPC moved the crane suspension point.

Controllers using nonlinear and energy-based control have also been studied for control crane systems. Early work considered two-dimensional overhead or gantry cranes. A nonlinear feedback controller was proposed in \cite{Yu1995} for a two-dimensional gantry crane where singular perturbation was used. This led to a composite controller with a slow tracking capability combined with fast oscillation damping. In \cite{Yoshida1998}, a two-dimensional gantry crane with constrained pendulum length and trolley motion was controlled using a Lyapunov-based nonlinear controller. LaSalle's invariance principle was used in \cite{Fang2001} to design a PD controller for tracking control of a two-dimensional overhead crane. This paper also included two nonlinear controllers based on PD control, where tracking and payload oscillation damping were improved by including nonlinear terms to account for coupling effects. An energy-based stabilizing feedback controller was presented in \cite{Sun2013} for a 4-DOF overhead crane for trolley position control and payload oscillation damping subject to input constraints. Nonlinear tracking control and swing damping of a three-dimensional overhead crane was proposed in \cite{Wu2016} using a feedback linearization approach. In \cite{Cibicik2018} an energy-based controller was proposed for damping the payload oscillations of a bifilar overhead crane. In \cite{Tysse2021}, a controller was proposed for a knuckle boom crane using vision-based feedback. The controller used an inner damping control loop to cancel payload oscillations and an outer PD controller to translate the crane suspension point. The vision system was used to estimate the payload oscillation angles and the crane cable length.

Learning-based and adaptive methods have also been applied to crane control to compensate for model uncertainties and disturbance forces. To deal with model uncertainties, \cite{Sun2015} proposed an energy-based adaptive controller for a planar overhead crane. The controller tracked the desired trolley position and cable length while damping payload oscillations and estimating the payload mass. In \cite{Sun2016}, an adaptive controller was proposed for automatic control of a tower crane under model uncertainties for position control and to limit payload oscillations.  An adaptive controller was proposed in \cite{Qian2017} where a learning algorithm was used to control a two-dimensional offshore boom crane subject to wave disturbances. The adaptive algorithm was used to compensate for disturbances by estimating unknown system parameters and the wave period. A neural network-based adaptive controller was proposed in \cite{Yang2020} for the control of a ship-mounted crane subject to ship roll motions and actuator input dead zones. The neural network was employed to approximate system uncertainties and dead-zone nonlinearities, while a sliding-surface design ensured convergence of boom and rope positioning. Experiments demonstrated robustness to parameter variations, external disturbances, and irregular wave-induced motions. \cite{Chen2021} addressed the control of a three-dimensional offshore rotary crane subject to ship roll disturbances. They incorporated the ship roll motion into the crane dynamics which simplified the subsequent controller design. An adaptive controller was developed to handle parametric uncertainties, and experimental validation demonstrated asymptotic stability and fast swing suppression, even under unmatched disturbances. In \cite{ZhangChen2022} they proposed an adaptive tracking control method for offshore cranes to estimate unknown gravity parameters which improved payload positioning accuracy and suppression of payload swinging from to persistent ship roll motion. The control of dual ship-mounted cranes was addressed in \cite{Qian2023} using an optimal learning sliding mode controller, leveraging a critic neural network to approximate the Hamilton–Jacobi–Bellman solution. The controller achieved precise positioning and swing suppression with robustness against parameter variations and external disturbances.

\subsection{Motivation}

Crane control systems have largely been based on accurate control of the suspension point in combination with damping of the payload pendulum motion, treating payload tracking and pendulum stabilization as separate problems (e.g. \cite{Tysse2021}). While this is natural with models based on Euler-angles, it can complicate integration with adaptive or learning-based controllers: If adaptation is introduced to counter external disturbances on the payload, there is a risk that the adaptive element compensates not only for external disturbances but also for the tracking action itself, thereby degrading payload tracking performance. MPC methods (e.g. \cite{Tysse2022}) achieve high tracking performance, but rely on accurate models and repeated online optimization, which increases the computational cost and limits scalability in fast or uncertain offshore conditions. Energy-based methods (e.g. \cite{Sun2013}) provide effective swing damping, but require precise system energy modeling and are less flexible under unmodeled disturbances such as varying wave excitations. Lastly, traditional adaptive methods (e.g. \cite{Sun2015}) generally estimate a small set of parametric uncertainties, which may be insufficient for highly variable offshore environments.

To address these issues, we propose a modeling approach that models the payload dynamics in Cartesian space rather than via Euler angles. Using the payload position as the primary controlled variable enables us to solve both the tracking and pendulum damping task with a unified framework, simplifying the resulting controller implementation and tuning. Such a formulation also enables the integration of learning-based methods to handle model uncertainties and disturbance rejection as the tracking controller and learning-based method act in concert, rather than as competing objectives. By combining this Cartesian model with the nonparametric adaptive control framework of \cite{Boffi2022}, disturbances are modeled in a reproducing kernel Hilbert space (RKHS). This allows adaptation to a broad class of unmodeled forces and uncertainties without requiring explicit parametric assumptions. To the best of our knowledge, this is the first application of nonparametric adaptive control to crane systems, made possible by the proposed Cartesian reformulation.

\subsection{Contribution}

In this paper we propose a payload tracking controller for an offshore crane using the novel nonparametric adaptive controller introduced in \cite{Boffi2022}. The controller models unmodeled dynamics and external disturbances as elements of an RKHS, allowing them to be learned directly from data. To apply this adaptive controller, the dynamic model used in \cite{Tysse2022} based on Euler angles is first reformulated to a model based on Cartesian coordinates, and then a tracking controller is designed using partial feedback linearization \cite{Spong1994}. This Cartesian model can also be used to achieve high-performance tracking in the nonadaptive case, as demonstrated in this paper. It is shown with Lyapunov-like analysis that the proposed controller gives uniformly ultimately bounded tracking errors and that the controller handles disturbances due to disturbance forces and unmodeled effects.

The contributions of the paper are:
\begin{enumerate}
    \item A new Cartesian model of the combined crane and payload dynamics is formulated by a change of coordinates from the usual model with Euler angles.
    \item A novel version of the partial feedback linearization method is formulated where the underactuated payload motion is controlled with the position of the crane tip instead of the acceleration of the crane tip. This is done by taking advantage of the gravity terms in the Cartesian model. 
    \item It is shown how to apply the nonparametric adaptive controller of \cite{Boffi2022} to the Cartesian model with partial feedback linearization, and stability properties are analyzed with and without the saturation function used in \cite{Boffi2022}.
    \item It is shown that partial feedback linearization with the Cartesian model improves payload tracking performance in the nonadaptive case compared to partial feedback linearization with the Euler angle model.
    \item The nonparametric adaptive controller is validated with good results in simulations and experiments where the base of the crane has a significant sinusoidal motion similar to a wave excitation on a ship. The disturbance was significantly reduced and the tracking performance was significantly improved with the adaptive compensation.
\end{enumerate}

\subsection{Paper structure}

This paper is organized as follows. The problem formulation and theoretical preliminaries are presented in Section \ref{sec:preliminaries}. The Cartesian model for the crane is developed in Section \ref{sec:modeling}. The control design is presented in Section \ref{sec:control_design}. The simulation studies and experimental validation are presented in Section \ref{sec:experiments}. Finally, the conclusion is presented in Section \ref{sec:conclusion}.
\section{Problem formulation}\label{sec:preliminaries}

\subsection{Crane model}

The nonparametric adaptive controller was applied to a novel dynamic model of the crane payload dynamics given in Cartesian coordinates. This model is given by
\begin{align}
    \ddot x + \omega_0^2 \frac{L_z}{L} x &= \omega_0^2 \frac{L_z}{L} x_0  + n_x + \sigma_x\label{eq:cartesian-intro-1}\\
    \ddot y + \omega_0^2 \frac{L_z}{L} y &= \omega_0^2 \frac{L_z}{L} y_0  + n_y + \sigma_y\label{eq:cartesian-intro-2}
\end{align}
Here $(x,y)$ is the horizontal position of the payload and  $(x_0,y_0)$ is the horizontal position of the suspension point at the crane tip, $L$ is the constant length of the crane cable, $L_z$ is the vertical distance from the crane tip to the cane load, $\omega_0 = \sqrt{g/L}$, $n_x$ and $n_y$ are higher order modeling terms and $\sigma_x$ and $\sigma_y$ are  generalized disturbance forces. This model structure was introduced since it allows for the use of the position $(x_0,y_0)$ of the suspension point as the control variable for the payload motion. This is done in a solution where the crane tip is controlled by a new formulation of partial feedback linearization \cite{Spong1994}. As will be shown in the following, this model structure is well suited for the application of the nonparametric adaptive controller of \cite{Boffi2022}.   In Section~\ref{sec:modeling} it is shown how the dynamic model (\ref{eq:cartesian-intro-1}, \ref{eq:cartesian-intro-2}) can be derived from the well-established dynamic model in Euler angles \cite{Tysse2022}:
\begin{align}
    \ddot\phi_x c_y + \omega_0^2 s_x &= \frac{c_x}{L}\ddot y_0  + n_{\phi_x} + \sigma_{\phi_x}\\
    \ddot\phi_y + \omega_0^2 c_x s_y &= -\frac{c_x}{L}\ddot x_0  + n_{\phi_y} + \sigma_{\phi_y}   
\end{align}
Here $\phi_x$ and $\phi_y$ are the Euler angles of the cable, $c_x = \cos\phi_x$, $s_x = \sin\phi_x$, $c_y = \cos\phi_y$, $s_y = \sin\phi_y$, $n_{\phi_x}$ and $n_{\phi_y}$ are higher order modeling terms and $\sigma_{\phi_x}$ and $\sigma_{\phi_y}$ are generalized disturbance forces. This model was used with partial feedback linearization in \cite{Tysse2022} where the crane tip was controlled with the acceleration $(\ddot x_0,\ddot y_0)$ of the crane tip to damp out the pendulum motion of the load. We found that this model in Euler angles was not straightforward to use with the nonparametric adaptive controller.

\subsection{Reproducing kernel Hilbert space}\label{sec:RKHS}

Methods based on reproducing kernel Hilbert spaces (RKHS) \cite{Aronszajn1950} are well established for data-driven identification of unknown functions \cite{Pillonetto2022}. In this paper we use the nonparametric adaptive controller proposed in \cite{Boffi2022}, where an RKHS formulation is used to approximate an unknown disturbance. A brief introduction to RKHS is presented in the following based on \cite{Micchelli2005} and \cite{Minh2011}. Let ${\boldK : \Rn \times \Rn \rightarrow \Rmm}$ be a matrix-valued reproducing kernel. Then the kernel will be positive definite in the sense that $    \boldK(\boldx,\boldz) = \boldK(\boldz,\boldx)^\T$ for all $\boldx, \boldz \in \Rn$ and 
$
    \sum_{i=1}^{N} \sum_{j=1}^{N} \innerproduct{\bolda_{i}}{\boldK(\boldx_{i},\boldx_{j})\bolda_{j}} \geq 0
$
for any sets of vectors ${\{ \boldx_i \}_{i=1}^{N} \in \Rn}$, ${\{\bolda_i \}_{i=1}^{N} \in \Rm}$ and for any integer $N>0$. Define the function ${\boldK_{x}\bolda : \Rn \rightarrow \Rm}$ by $\boldK_{x}\bolda= \boldK(\cdot,\boldx)\bolda$ which gives
$
    (\boldK_{x}\bolda)(\boldz) = \boldK(\boldz,\boldx)\bolda \in \Rm
$
for all $\bolda \in \Rm$ and $\boldx,\boldz \in \Rn$. Then the reproducing kernel $\boldK$ defines the RKHS ${\RKHS}$ given by
\begin{equation}
   \RKHS = \overline{\spn}\{ \boldK_{x}\bolda \; | \; \boldx \in \Rn, \bolda \in \Rm\}
\end{equation}
The reproducing kernel function can be expressed in terms of a feature map $\boldPhi(\boldx)$ as $\boldK(\boldx,\boldz) = \boldPhi(\boldx)^\T\boldPhi(\boldz)$. 

Suppose that the kernel $\boldK$ is shift invariant. Then from the vector version of Brochner's theorem  there is a matrix function $\boldM:\R^n\rightarrow\CC^{m\times m}$ and a probability density function $p(\boldw)$ for $\wv\in\R^n$ so that \cite{Brault2016} 
\begin{align}
    \boldK(\boldx,\boldz) 
    &= 
    \int_{\R^n} \boldPhi(\boldx,\boldw)^* \boldPhi(\boldz,\boldw) p(\boldw) d\boldw
    \label{Brochners-theorem}
\end{align}
where the RFF feature map $\boldPhi(\boldx,\boldw) \in\CC^{2m\times m}$ is given by
\begin{equation}
    \label{general-RFF-feature-map}
   \boldPhi(\boldx,\boldw) 
   =  \bb
        \cos(\boldw^{\T} \boldx)\boldM(\boldw)^*\\
        \sin(\boldw^{\T} \boldx)\boldM(\boldw)^*
        \eb  
\end{equation}
A function $\boldh\in \RKHS$ can then be written as 
\begin{align}
    \boldh &= 
    \int_{\R^n} 
    \boldPhi(\cdot,\boldw)^*\boldalpha(\boldw) p(\boldw) d\boldw
    \in\RKHS
    \label{h-in-F2}
\end{align}
where $\boldalpha(\wv) = \sum_{j=1}^{\infty} \boldPhi(\boldx_i,\boldw) \bolda_i \in \CC^{2m}$. 
A random Fourier feature (RFF) approximation is given by \cite{Brault2016,Rahimi2007}
\begin{align}
    \boldh &\approx \frac{1}{d}
    \sum_{i=1}^{d} \boldPhi(\cdot,\boldw_i)^*\boldalpha(\wv_i) 
    \label{RFF-h-approx}
\end{align}
where $\boldw_1,\ldots,\boldw_d$ are drawn i.i.d. with probability $p(\boldw)$. The number of random features ${d}$ is chosen to balance the quality of the approximation with respect to the computational requirements. The notation 
\begin{equation}
    \frac{1}{d}
    \sum_{i=1}^{d} \boldPhi(\boldx,\boldw_i)^* \boldalpha_i = \boldPsi(\boldx)^* \boldalpha 
    \label{RFF-h-approx_Psi}
\end{equation}
is used where $\boldalpha = [\boldalpha_1^\T,\ldots,\boldalpha_{d}^\T]^\T\in \CC^{2dm}$ and $\boldalpha_i = \boldalpha(\wv_i)$. The RFF feature map $\boldPsi(\boldx)\in \CC^{2dm \times m}$ is \cite{Sindhwani2018,Singh2021}
\begin{equation}\label{RFF-def-Psi}
    \boldPsi(\boldx) 
    = \frac{1}{d}
    \bb
        \cos(\boldw_{1}^{\T} \boldx)\boldM(\boldw_1)^*\\
        \sin(\boldw_{1}^{\T} \boldx)\boldM(\boldw_1)^*\\
        \vdots\\
        \cos(\boldw_{d}^{\T} \boldx)\boldM(\boldw_d)^*\\
        \sin(\boldw_{d}^{\T} \boldx)\boldM(\boldw_d)^*
    \eb  
\end{equation}

\subsection{Gaussian separable kernel}
\label{sec:gaussian-kernel}

In this paper, the Gaussian separable kernel \cite{Sindhwani2018} will be used. This is the shift-invariant reproducing kernel
$
    \boldK(\boldx,\boldz) = k(\boldx,\boldz) \eye_m
    \in \Rmm
$
where $\eye_m$ is the $m\times m$ identity matrix and
$
    k(\boldx,\boldz) = \exp\left( -\|\boldx - \boldz \|_2^{2}/(2 \sigma^{2}) \right)
$
is the scalar shift-invariant Gaussian kernel with kernel width ${\sigma > 0}$.
For the Gaussian separable kernel \eqref{Brochners-theorem} holds with $\boldM(\boldw) = \eye_m$ and $p(\boldw) = \mathcal{N}(\zero,\sigma^{-2}\eye_n)$ and $\boldPhi(\boldx,\boldw) = [\cos(\boldw^\T\boldx),\sin(\boldw^\T\boldx)]^\T$. The RFF feature map \eqref{general-RFF-feature-map} can then be written as  
\begin{equation}
   \boldPhi(\boldx,\boldw) 
   =  \bb
        \cos(\boldw^{\T} \boldx)\eye_m\\
        \sin(\boldw^{\T} \boldx)\eye_m
        \eb  
        \in\R^{2m\times m}
        \label{Gaussian-RFF-feature-maps}
\end{equation}
and $\boldPsi(\boldx)$ is given by \eqref{RFF-def-Psi} with $\boldM(\boldw) = \eye_m$.  

\begin{assumption}\label{ass-Gaussian-kernel}
    Let the RKHS $\RKHS$ be defined by the Gaussian separable kernel $\boldK(\boldx,\boldz)$ with RKK feature map $\boldPhi(\boldx,\boldw)\in\R^{2m\times m}$ given by \eqref{Gaussian-RFF-feature-maps}. Let ${\boldw_{1}, \dots, \boldw_{d} \in \Rn}$ be i.i.d. with distribution $p(\wv)={\calN\left(\zero, \sigma^{-2} \eye_n \right)}$. The function $\boldh\in\RKHS$ is given by \eqref{h-in-F2} where $\boldalpha(\wv)\in\R^{2m}$ and $\sup_{\wv\in\R^n}\|\boldalpha(\wv)\|_2\leq B_h$ for some constant $B_h>0$.  
\end{assumption}

\begin{proposition}
    Suppose that Assumption \ref{ass-Gaussian-kernel} holds. Then the RFF feature map \eqref{Gaussian-RFF-feature-maps} has operator norm $\|\boldPhi(\boldx,\boldw)\|_2=1$, and the RFF feature map $\boldPsi(\boldx)$ defined in \eqref{RFF-def-Psi} is locally Lipschitz. 
    \label{prop:Gaussian_bound-Lipschitz}
\end{proposition}
{\it Proof:} This follows from $\|\boldPhi\|_2^2 = \|\boldPhi^\T \boldPhi\|_2$, and since sine and cosine are locally Lipschitz. \hboks

\subsection{Bound on RFF approximation error}

A bound on the approximation error in \eqref{RFF-h-approx} is given in this section. This is based on the work of  \cite{Rahimi2008} for the scalar case and \cite{Boffi2022} for the vector case. 

\begin{assumption}\label{ass:x-and-alpha-bounds}
    Let $X\subset\R^n$ be a compact set and let $B_X = \sup_{x\in X}\|\boldx\|_2$. Let $\|\f(\cdot)\|_\infty=\sup_{x\in X}\|\f(\boldx)\|_2$.
\end{assumption}

\begin{proposition}\label{prop:Gaussian-RFF-bound}
    Let Assumptions \ref{ass-Gaussian-kernel} and \ref{ass:x-and-alpha-bounds} hold. Fix $\delta \in (0,1)$, $B_h>0$ and a positive integer $d$. Then, with probability $1-\delta$ there exist weights $\boldalpha_i\in\R^{2m}$ so that $\|\boldalpha_i\|_2\leq B_h$ for $i=1,\ldots,d$ and
\begin{equation}
\left\| \frac{1}{d}
\sum_{i=1}^{d} \boldPhi(\cdot,\boldw_i)^\T \boldalpha_i - \boldh(\cdot)\right\|_\infty \leq B_\epsilon
\end{equation}
where 
\begin{equation}
        B_\epsilon = \frac{4B_h}{\sqrt{d}}\left( \frac{B_X\sqrt{n}}{\sigma} + \sqrt{m} + g(\delta) \right)
        \label{B-epsilon-def}
\end{equation}
and $g(\delta) = \frach\left(\sqrt{\log(2/\delta)} + \sqrt{\delta/2}\right)$. 
 \end{proposition}
The proof follows the proofs of \cite[Propositions 5.1 and 5.2]{Boffi2022} closely and is not included here. The difference is that the RFF feature map that we use gives $\|\boldPhi(\boldx,\boldw)\|_2=1$ and $\Phim(\x,\wv_i)^\tr\boldalpha_i = c_i\boldalpha_{i,c}+s_i\boldalpha_{i,s}$ where $c_i=\cos(\wv_i^\tr\x)$, $s_i=\sin(\wv_i^\tr\x)$, $\boldalpha_i = [\boldalpha_{i,c}^\T,\boldalpha_{i,s}^\T]^\T$. Then $\|\boldalpha_{i,c}c_i+\boldalpha_{i,s}s_i\|^2\leq\|\boldalpha_{i,c}\|^2c_i^2+\|\boldalpha_{i,s}\|^2s_i^2\leq \|\boldalpha_i\|^2$. Moreover, $\EE\left[\|\wv_i\|_2^2\right]=n\sigma^{-2}$ for the Gaussian separable kernel. The Rademacher complexity bound is then found from the last equation in \cite[Appendix~D.2]{Boffi2022} to be
$
\EE \left[\left\| \sum_{i=1}^{d} \varepsilon_i\boldPhi (\cdot,\wv_i)\boldalpha_i \right\|_\infty \right]
\leq 2\sqrt{{d}} B_h 
\left[\frac{B_X\sqrt{n}}{\sigma} + \sqrt{m} \right]
$ 
where $\varepsilon_1,\ldots,\varepsilon_d$ are Rade\-macher random variables. The truncation of $\boldPhi(\boldx,\boldw)$ used in \cite{Boffi2022} is not necessary since $\|\boldPhi(\boldx,\boldw)\|_2=1$ by Proposition \ref{prop:Gaussian_bound-Lipschitz}.



\subsection{Nonparametric adaptive controller}

In this paper, the nonparametric adaptive controller of \cite{Boffi2022} is used for tracking control of the crane payload. This controller was developed in \cite{Boffi2022} for the nonlinear system 
\begin{equation}
    \boldxdot = \boldf(\boldx,t) + \boldB(\boldx,t) (\boldu(\boldx,t) - \boldh(\boldx))
    \label{general-nonlinear-dynamics}
\end{equation}
where ${\boldx \in X\subset \Rn}$ is the system state, ${t \in  \R_{\geq 0}}$ is time, $\boldf : \Rn \times \R_{\geq 0} \rightarrow \Rn$ are the nominal dynamics, $\boldB : \Rn \times \R_{\geq 0} \rightarrow \R^{n \times m}$ is the control matrix, ${\boldu : \Rn \times \R_{\geq 0} \rightarrow \Rm}$ is the learned control input, and $\boldh:\Rn \rightarrow \Rm$ is the unknown disturbance term, which is assumed to be an element of the RKHS $\RKHS$, which is defined by the reproducing kernel $\boldK :\Rn \times \Rn \rightarrow \Rmm$ with a feature map $\boldPhi(\boldx)$ which satisfies $\boldK(\boldx,\boldz) = \boldPhi(\boldx)^\T\boldPhi(\boldz)$. It is noted that the feature map $\boldPhi(\boldx)$ is infinite-dimensional for the Gaussian kernel.  This is shown in Appendix~\ref{sec:appendix_gaussian_kernel}.

The tracking error ${\bolde \in \Rn}$ is defined as ${\bolde = \boldx - \boldx_d}$ where ${\boldx_d \in \Rn}$ is the desired trajectory. The error dynamics are assumed to be 
uniformly asymptotically stable and given by
\begin{equation}
\label{error-dynamics-problem-formulation}
    \boldedot = \boldf_e(\bolde,t) + \boldB(\boldx,t) (\boldu(\boldx,t) - \boldh(\boldx))
\end{equation}

An adaptive control law that compensates for the unknown disturbance $\boldh(\boldx)$ in \eqref{error-dynamics-problem-formulation} is given by 
\begin{equation}
    \label{u=hat-h}
    \boldu(\boldx,t) = \hat{\boldh}(\boldx,t)
\end{equation}
where $\hat{\boldh}(\cdot,t)$ is an estimate of $\boldh\in\RKHS$. When a Lyapunov function $Q(\bolde,t)$ is given for the nominal error dynamics it is a well-established approach \cite[page 5]{Boffi2021} to use an adaptive control law of the form 
\begin{align}
   \hat{\boldh}(\boldx,t) &= \boldY(\boldx)^{\T} \hat{\boldbeta}(t) 
   \label{adaptive-sanner-h-hat}
   \\
    \Dot{\hat{\boldbeta}} &= - \gamma \boldY(\boldx) \boldB(\boldx,t)^{\T} \nabla Q(\bolde,t)
    \label{adaptive-sanner-update-beta-hat}
\end{align}
where $\boldY(\boldx)$ is a matrix of known basis functions. The estimate $\hat{\boldh}$ is then the linear combination of a finite number of given basis functions. Such basis functions can be model-based as in the adaptive robot tracking controller of \cite{Slotine1987} where uncertain terms are structurally known and only parameter values must be determined. Another possible solution is to us Gaussian basis functions placed in a fixed mesh arrangement as in \cite{Sanner1992} where the parameter estimate will be in the span of the basis functions. 

A different approach, which is used in this paper, is the nonparametric adaptive controller of \cite{Boffi2022} where $\boldY(\boldx)$ is set to the infinite-dimensional feature map $\boldPhi(\boldx)$ of a reproducing kernel $\boldK(\boldx,\boldz) = \boldPhi(\boldx)^\T\boldPhi(\boldz)$. The estimate $\hat{\boldh}$ is then found as an element of the infinite-dimensional RKHS $\RKHS$. This nonparametric adaptive controller is given by 
\begin{align}
    \hat{\boldh}(\boldx,t) &= \int_0^t \boldK(\boldx,\boldx(\tau)) \boldc(\tau) d\tau
    \label{nonparametric-hat-h-kernel}\\
    \boldc(t) &= - \gamma \boldB(\boldx,t)^{\T} \nabla Q(\bolde,t)
    \label{nonparametric-c(t)}
\end{align}
which is a reformulation of (\ref{adaptive-sanner-h-hat}, \ref{adaptive-sanner-update-beta-hat}) based on the RKHS kernel trick, where the kernel of dimension $m\times m$ is used instead of the infinite-dimensional feature map. The equivalence of (\ref{nonparametric-hat-h-kernel}, \ref{nonparametric-c(t)}) and (\ref{adaptive-sanner-h-hat}, \ref{adaptive-sanner-update-beta-hat}) is verified by letting $\boldY = \boldPhi$ and noting that the time integral of \eqref{adaptive-sanner-update-beta-hat} is 
\[
\hat{\boldbeta}(t) = - \gamma\int_0^t \boldPhi(\x(\tau))  \boldB(\x(\tau),\tau)^\T\nabla Q(\bolde(\tau),\tau) d\tau
\]
When the separable Gaussian kernel from Section~\ref{sec:gaussian-kernel} is used, the estimate becomes
\begin{align}
    \hat{\boldh}(\boldx,t) &= \int_0^t \exp{\left( -\frac{\| \boldx - \boldx(\boldtau) \|^{2}}{2 \sigma^{2}} \right)} \boldc(\tau) d\tau
    \label{nonparametric-gaussian-hat-h-kernel}
\end{align}
This shows the data-driven nature of the estimate \eqref{nonparametric-hat-h-kernel} where the estimate is given as a weighted integral of Gaussian functions along the system trajectory.

Notable features of the proposed controller (\ref{nonparametric-hat-h-kernel}, \ref{nonparametric-c(t)}) is that 
$\hat{\boldh}\in\RKHS$, since $\boldK(\cdot,\boldx(\tau))\boldc(\tau)=\boldK_{\boldx(\tau)} \boldc(\tau)\in \RKHS$. Moreover, the basis functions are data-driven, and the application of the kernel trick makes it possible to use $\boldY = \boldPhi$ and a parameter vector $\hat\boldbeta$ of infinite dimension since $\boldPhi$ and $\hat\boldbeta$ do not appear in (\ref{nonparametric-hat-h-kernel}, \ref{nonparametric-c(t)}), instead, only the kernel $\boldK(\boldx,\boldx(\boldtau)) = \boldPhi(\boldx)^\T\boldPhi(\boldx((\boldtau))$ is used.

\begin{assumption}
    \label{ass-system-and-error-regularity}
    The functions $\boldf(\boldx,t)$ and $\B(\boldx,t)$ are known, and $\boldf(\boldx,t)$, $\B(\boldx,t)$ and $\boldh$ are locally Lipschitz in $\boldx$ and locally bounded in $\boldx$ uniformly in $t$. The error is $\bolde = \boldx - \boldx_d$, and the function $\boldf_e(\bolde,t)$ is locally Lipschitz in $\bolde$ and locally bounded in $\bolde$ uniformly in $t$.  
\end{assumption}

\begin{assumption}
    \label{Lyapunov-function-nominal-error-dynamics}
    There is a Lyapunov function $Q(\bolde)$ for the error system \eqref{error-dynamics-problem-formulation} so that $\nabla Q(\bolde,t)$ and $\partial Q(\bolde,t)/\partial t$
are locally bounded in $\bolde$ uniformly in $t$, $\nabla Q(\bolde,t)$ is locally Lipschitz in $\bolde$ and 
\begin{align}
    \nabla Q(\bolde,t)^\T\boldf_e(\bolde,t) + \frac{\partial Q}{\partial t} \leq -\rho(\|\bolde\|_2)
\\
    \mu_1(\|\bolde\|_2)
    \leq
    Q(\bolde,t)
    \leq \mu_2(\|\bolde\|_2)
\end{align}
where $\rho$, $\mu_1$ and $\mu_2$ are class $\mathcal K_\infty$ functions \cite[page 144]{Khalil2002}.
\end{assumption}

\begin{theorem}
    Consider the system \eqref{general-nonlinear-dynamics}
 under Assumptions \ref{ass-Gaussian-kernel}, \ref{ass:x-and-alpha-bounds}, \ref{ass-system-and-error-regularity} and \ref{Lyapunov-function-nominal-error-dynamics}. Let $\gamma>0$. Then the adaptive control law $\boldu(\boldx,t) = \hat{\boldh}(\boldx,t)$ where $\hat{\boldh}(\boldx,t)$ is given by  (\ref{nonparametric-hat-h-kernel}, \ref{nonparametric-c(t)}) will ensure that $\boldx(t)$ and $\bolde(t)$ exist and are uniformly bounded for all $t\geq 0$, $\boldu\in\RKHS$ and $\lim_{t\rightarrow\infty}\|\bolde(t)\| = 0$.
 \end{theorem}
 The proof is a special case of the proof of \cite[Theorem 4.5]{Boffi2022}. 

\subsection{RFF approximation of adaptive control law}

The computational requirements of the adaptive control law (\ref{nonparametric-hat-h-kernel}, \ref{nonparametric-c(t)}) do not allow for real-time computation. This problem was solved in \cite{Boffi2022} where the function $\boldh\in\RKHS$ is approximated by the RFF approximation given by \eqref{RFF-h-approx}. This gives 
\begin{equation}
    \boldh(\boldx) = \boldPsi(\boldx)^{\T} \alphav + \epsv(\boldx)
    \label{h=Psi-alpha+epsilon}
\end{equation}
where it follows from Proposition~\ref{prop:Gaussian-RFF-bound} that the approximation error $\epsv(\boldx)\in\R^m$ is bounded by 
\begin{equation}
    \|\epsv\|_\infty \leq B_\epsilon
    \label{epsilon-bound}
\end{equation} 
where $B_\epsilon$ is given by \eqref{B-epsilon-def}. It is noted that the bound $B_\epsilon$ on the approximation error can be made arbitrarily small by increasing the number of $d$ of RFF features, and $\|\boldepsilon\|_\infty\rightarrow 0$ when $d\rightarrow \infty$. 
The RFF approximation of the nonparametric adaptive control law (\ref{nonparametric-hat-h-kernel}, \ref{nonparametric-c(t)}) is then given by 
\begin{align}
   \hat{\boldh}(\boldx,t) &= \boldPsi(\boldx)^{\T} \hat{\boldalpha}(t) 
   \label{adaptive-nonparametric-h-hat}
   \\
    \Dot{\hat{\boldalpha}} &= - \gamma \boldPsi(\boldx) \boldB(\boldx,t)^{\T} \nabla Q(\bolde,t)
    \label{adaptive-nonparametric-update-beta-hat}
\end{align}
This formulation has the same structure as the one in (\ref{adaptive-sanner-h-hat}, \ref{adaptive-sanner-update-beta-hat}). The difference is that the controller (\ref{adaptive-nonparametric-h-hat}, \ref{adaptive-nonparametric-update-beta-hat}) is an approximation of the infinite-dimensional adaptive control law (\ref{nonparametric-hat-h-kernel}, \ref{nonparametric-c(t)}), where the approximation error $\boldepsilon$ is bounded. 
The motivation for using the approximation (\ref{adaptive-nonparametric-h-hat}, \ref{adaptive-nonparametric-update-beta-hat}) is to have a controller that can be computed in real time.  It is noted that the estimation error of the nonparametric adaptive controller is
\begin{equation}
    \hat\boldh(\boldx) - \boldh(\boldx) = \boldPsi(\boldx)^{\T} \alphavt - \epsv(\boldx)
    \label{nonparametric-estimation error}
\end{equation}
where $\alphavt = \alphavh - \alphav$ is the parameter estimation error.



\section{Modeling}\label{sec:modeling}

\subsection{Payload dynamics in Euler angles}

The crane is modeled as a spherical pendulum with moving suspension point as shown in Figure~\ref{fig:figure_crane_modeling}. Let ${n}$ be the inertial frame with origin at the base of the crane and the ${z}$-axis vertically up. Let ${b}$ be the moving frame with origin at $\boldr_0 = [x_0,y_0,z_0]^\tr$, which is the position of the suspension point of the cable in the $n$ frame, and with the ${z}$-axis along the cable. The rotation from frame ${n}$ to frame ${b}$ is given by the rotation matrix ${\boldR_{b}^{n} = \boldR_x(\phi_x) \boldR_y(\phi_y)}$ where $\boldR_x$ and $\boldR_y$ are the rotation matrices about the $x$ and $y$ axes and $\phi_x$ and $\phi_y$ are the angles of rotation \cite{Siciliano2008}. The position of the load mass in the $n$ frame is $\rv = \rv_0 - \boldR^n_b[0,0,L]^\tr$ with coordinates $\boldr = [x,y,z]^\tr$, and the relative position of the mass with respect to the crane tip is $\boldr_r =\boldr-\boldr_0=- \boldR^n_b[0,0,L]^\tr$ with coordinates $\boldr_r = [x_r,y_r,z_r]^\T$. The constant length of the massless cable is
$
    L = \sqrt{x_r^2 + y_r^2 + z_r^2}
$.
It is assumed that the suspension point moves in the horizontal plane. The equations of motion for the load mass are derived with Kane's equations of motion in \cite[eq.~(7)]{Tysse2022} and are given by
\begin{align}
     \phiddot_x c_y + \omega_{0}^{2} s_x 
     &= - \frac{1}{L} \yddot_0 c_x + 2 \phidot_x \phidot_y s_y + \frac{c_x}{mL}F_y
\label{eq:equations_of_motion_using_omega_1}\\
     \phiddot_y + \omega_{0}^{2} c_x s_y 
     &= \frac{1}{L} \xddot_0 c_y + \frac{1}{L} \yddot_0 s_x s_y - \phidot_x^2 s_y c_y\nonumber\\
     & \quad - \frac{c_y}{mL}F_x  - \frac{s_xs_y}{mL}F_y
     \label{eq:equations_of_motion_using_omega_2}
\end{align}
The pendulum motion can then be controlled with the accelerations $(\ddot x_0,\ddot y_0)$ of the suspension point as in \cite{Tysse2022}. 
\begin{figure}[t]
    \centering
    \input{figures/models/figure_crane_modeling}
    \caption{Model of the crane system showing the payload mass $m$ with position $\boldr = [x,y,z]^{\T}$ connected to the suspension point with position $\boldr_0 = [x_0,y_0,z_0]^{\T}$ by a cable with length $L$. The inertial frame $n$ is centered at the base of the crane, and moving frame $b$ centered at the suspension point with $z$-axis along the cable.}
    \label{fig:figure_crane_modeling}
\end{figure}

\subsection{Payload dynamics in Cartesian coordinates}
\label{sec:spherical_pendulum_moving_pivot_cartesian_change_coordinate}

In this section a Cartesian model is derived. The relative positions are given by
\begin{equation}
\label{Change-of-Coordinates-1}
    [x_r, y_r, z_r]^\tr = [- s_y L,s_x c_y L,- c_x c_y L]^\tr
\end{equation}
It is assumed that $z_r < 0$, which means that the load is below the suspension point. Then
\begin{equation}
\label{Change-of-Coordinates-Lz}
    L_z = - z_r = \sqrt{L^2 - x_r^2 - y_r^2} \geq 0
\end{equation}
The relative horizontal velocities are then given by
$\xdot_r = - \phidot_y c_y L$ and $\ydot_r =   \phidot_x c_x c_y L - \phidot_y s_x s_y L$ 
while the relative horizontal accelerations are
\begin{align}
\label{Change-of-Coordinates-3}
    \xddot_r 
    &=  - \phiddot_y c_y L + \phidot_y^2 s_y L\\
    \yddot_r 
    &=  \phiddot_x c_x c_y L - \phiddot_y s_x s_y L - \phidot_x^2 s_x c_y L\nonumber\\
    &\quad- \phidot_y^2 s_x c_y L - 2 \phidot_x \phidot_y c_x s_y L
    \label{Change-of-Coordinates-ddot_yr}
\end{align}
The equations of motion in the Cartesian coordinates $(x,y)$ are then found by solving for $\ddot\phi_x$, $\ddot\phi_y$, $\dot\phi_x$, $\dot\phi_y$, $c_x$, $s_x$, $c_y$ and $s_y$ from (\ref{Change-of-Coordinates-1}--\ref{Change-of-Coordinates-ddot_yr}) and inserting the expressions into the equations of motion (\ref{eq:equations_of_motion_using_omega_1}, \ref{eq:equations_of_motion_using_omega_2}). A detailed derivation of the Cartesian model is presented in Appendix~\ref{sec:appendix_cartesian_modeling}. This gives the model 
\begin{align}
\label{eq:cartesian-modeling-1}
\ddot x + \Omega_z^2 x
&= \Omega_z^2 x_0 + n_{ax} + n_{vx} + \sigma_x \\
\label{eq:cartesian-modeling-2}
\ddot y + \Omega_z^2 y
&=  \Omega_z^2 y_0 + n_{ay} + n_{vy} + \sigma_y
\end{align}
where $\Omega_z^2 = \omega_0^2\frac{L_z}{L} \leq \omega_0^2$. The acceleration terms are
\begin{align}
\label{eq:equations_of_motion_n_ax}
        n_{ax} &= \frac{x_r^2}{L^2} \xddot_0 + \frac{x_ry_r}{L^2} \yddot_0\\
        \label{eq:equations_of_motion_n_ay}
        n_{ay} &=   \frac{x_ry_r}{L^2}\xddot_0 + \frac{y_r^2}{L^2}\yddot_0
\end{align}
The velocity-related terms are 
\begin{align}
\label{eq:equations_of_motion_using_x_m_n_x}
    \begin{split}
        n_{vx} &= -\frac{x_r\xdot_r^2}{L^2 - x_r^2}
        - \frac{x_r^3 y_r^2\xdot_r^2}{L^2L_z^2(L^2 - x_r^2)}\\
        &\quad - 2 \frac{x_r^2y_r\xdot_r\ydot_r}{L^2L_z^2}
        - \frac{x_r(L^2 - x_r^2)\ydot_r^2}{L^2L_z^2}
    \end{split}\\
\label{eq:equations_of_motion_using_y_m_n_y}
    \begin{split}
        n_{vy} &=  -\frac{y_r\xdot_r^2}{L^2 - x_r^2} - \frac{x_r^2 y_r^3 \xdot_r^2} {L^2L_z^2(L^2 - x_r^2)}\\
        &\quad - 2 \frac{x_ry_r^2 \xdot_r\ydot_r}{L^2L_z^2} - \frac{y_r(L^2 - x_r^2) \ydot_r^2}{L^2L_z^2}
    \end{split}
\end{align}
and disturbance forces $F_x$ and $F_y$ in the $x$ and $y$ directions of the $n$ frame result in the terms
\begin{align}
\label{eq:equations_of_motion_sigma_x}
        \sigma_x &= \frac{y_r^2+z_r^2}{mL^2}F_x - \frac{x_ry_r}{mL^2}F_y\\
        \label{eq:equations_of_motion_sigma_y}
        \sigma_y &=  -\frac{x_ry_r}{mL^2}F_x + \frac{x_r^2 + z_r^2}{mL^2}F_y
\end{align}

The equations of motion (\ref{eq:cartesian-modeling-1}, \ref{eq:cartesian-modeling-2}) can be controlled with the position $(x_0,y_0)$ of the suspension point. The equations (\ref{eq:cartesian-modeling-1}, \ref{eq:cartesian-modeling-2}) have more terms and appear to be more complicated than the equations of motion (\ref{eq:equations_of_motion_using_omega_1}, \ref{eq:equations_of_motion_using_omega_2}) in Euler angles. However, all terms in $n_{ax}$, $n_{ay}$, $n_{vx}$ and $n_{vy}$ are higher order terms that can be treated as vanishing perturbations in a controller design where the nominal dynamics are exponentially stable \cite{Khalil2002}. Therefore, these terms are handled without much complication in the controller design used in this paper. 
\section{Control design}\label{sec:control_design}

\subsection{Partial feedback linearization}

A novel method for partial feedback linearization \cite{Spong1994} is presented in this section for the system consisting of the actuated crane and the unactuated crane load. The new idea that we propose is to use the position of the actuated part to control the unactuated dynamics by taking advantage of the gravity terms in the model. This is different from the original method of partial feedback linearization of \cite{Spong1994}, which was used in \cite{Tysse2022}, where the acceleration of the actuated part was used to control the unactuated part. The new formulation is made possible by formulating the load model in Cartesian coordinates, instead of the Euler angle model used in \cite{Tysse2022}. 

The generalized coordinates of the crane and the load are $\boldq = [\boldq_1^\T,\boldq_2^\T]^\T$ where $\boldq_1 = [\phi_x,\phi_y]^\T$ are the Euler angles of the load and $\boldq_2 = [q_1,q_2,q_3]^\T$ are the joint angles of the crane. The corresponding input generalized forces of the crane are $\boldtau_q =[\tau_1,\tau_2,\tau_3]^\T$. The dynamics are given by the underactuated system
\begin{align}
    \boldM_{11}\boldqddot_1 + \boldM_{12} \boldqddot_2 + \boldc_{q1} + \boldg_{q1} &= \boldsigma_q \label{eq:dynamics_unactuated}\\
    \boldM_{21}\boldqddot_1 + \boldM_{22} \boldqddot_2 + \boldc_{q2} + \boldg_{q2} &= \boldtau_q \label{eq:dynamics_actuated}
\end{align}
where $\boldc_{q1} = \boldC_1(\boldq,\boldqdot)\boldqdot$ and $\boldc_{q2} = \boldC_2(\boldq,\boldqdot)\boldqdot$ are centrifugal and Coriolis terms, and $\boldg_{q1}$ and $\boldg_{q2}$ are gravitational terms. The term $\boldsigma_q$ is an unknown generalized disturbance force acting on the load. The mass matrix 
\beq
\M(\boldq) = \ba{cc} \M_{11}(\boldq) & \M_{12}(\boldq) \\ \M_{21}(\boldq) & \M_{22}(\boldq) \ea 
\eeq
is symmetric and positive definite with elements $M_{ij}(\boldq)$ and satisfies the properties of a mass matrix with revolute joints as given in \cite[page 96]{Kelly2005}. Moreover, $\|\boldc_{q1}\| \leq C_{cq1}\|\dot\q\|^2$ where $C_{cq1} > 0$ is a constant \cite[page 99]{Kelly2005}. Here \eqref{eq:dynamics_unactuated} is a reformulation of (\ref{eq:equations_of_motion_using_omega_1}, \ref{eq:equations_of_motion_using_omega_2}), while \eqref{eq:dynamics_actuated} is found as standard manipulator dynamics \cite{Siciliano2008}. 
A change of variables to $\p = [\yv^\tr,\yv_0^\tr]^\tr$ where $\boldy = [x,y]^\T$ and $\boldy_0 = [x_0,y_0]^\T$ is done. The velocity mappings are $ \boldydot = \boldJ_1(\boldq_1) \boldqdot_1$ and $ \boldydot_0 = \boldJ_2(\boldq_2) \boldqdot_2$. The dynamics are then 
\begin{align}
    \boldD_{11}\boldyddot + \boldD_{12} \boldyddot_0 + \boldc_1 + \boldg_1 &= \boldsigma \label{eq:dynamics_unactuated_Y}\\
    \boldD_{21}\boldyddot + \boldD_{22} \boldyddot_0 + \boldc_2 + \boldg_2 &= \boldtau \label{eq:dynamics_actuated_Y}
\end{align}
where $\boldsigma=\boldJ_1(\boldq_1)^{-\T}\boldsigma_q$, $\boldtau = \boldJ_2(\boldq_2)^{-\T}\boldtau_q$ and 
\beq\nonumber
\Dm = \ba{cc} \Dm_{11} & \Dm_{12} \\ \Dm_{21} & \Dm_{22} \ea 
= \bb 
\boldJ_1^{-\T}\boldM_{11}\boldJ_1^{-1} & \boldJ_1^{-\T}\boldM_{12}\boldJ_2^{-1} \\ 
\boldJ_2^{-\T}\boldM_{21}\boldJ_1^{-1} & \boldJ_2^{-\T}\boldM_{22}\boldJ_2^{-1} 
\eb 
\eeq
The Jacobians $\boldJ_1$ and $\boldJ_2$ and their inverses are assumed to be bounded, which is a reasonable assumption for a crane. The positive definite mass matrix $\Dm$ then satisfy the properties of a mass matrix with revolute joints as given in \cite[page 96]{Kelly2005}. In particular, the induced 2-norm of $\boldD$ is upper and lower bounded by $\alpha_1 \leq \|\boldD\|_2 \leq \alpha_2$ for some $\alpha_2>\alpha_1>0$. 

Equation \eqref{eq:dynamics_unactuated_Y} is a reformulation of (\ref{eq:cartesian-modeling-1}, \ref{eq:cartesian-modeling-2}), which means that $\boldD_{11} = \eye$, $\boldc_1 = [n_{vx},n_{vy}]^\T$, $\boldsigma = [\sigma_x,\sigma_y]^\T$,
\begin{equation}
\label{g1-expression}
    \boldg_1 =  \Omega_z^2(\boldy - \boldy_0)
    = \Omega_z^2[ x_r, y_r]^\T
\end{equation}
and
\begin{equation}
    \boldD_{12} = \boldD_{21} = 
    \bb \frac{x_r^2}{L^2} & \frac{x_ry_r}{L^2} \\
    \frac{x_ry_r}{L^2} & \frac{y_r^2}{L^2} 
    \eb
\end{equation}
Then $|x_r|\leq L$ and $|y_r|\leq L$ implies that $\|\boldg_1\|\leq \omega_0^2 L$ and that $\boldD_{21}$ is bounded with finite induced norm $\|\boldD_{21}\|_2\leq B_{12}$ for some $B_{12}>0$.
Moreover, since $\boldc_1 = [n_{vx},n_{vy}]^\T$ it is seen from \eqref{eq:equations_of_motion_n_ax} and \eqref{eq:equations_of_motion_n_ay} that 
$
    \|\boldc_1\|_2 \leq C_{c1}\|\dot\p\|_2^2 
    = C_{c1}\left(\|\dot\yv\|_2^2 + \|\dot\yv_0\|_2^2\right) 
$
where $C_{c1} > 0$ is a constant.  The expression $\ddot \yv = -\boldD_{12} \boldyddot_0 - \boldc_1 - \boldg_1 + \boldsigma$ is found from \eqref{eq:dynamics_unactuated_Y} with $\boldD_{11}=\eye$, and insertion into \eqref{eq:dynamics_actuated_Y} gives
\begin{equation}
\label{eq:M22bar_dynamics_QY}
    \boldDbar_{22} \boldyddot_0 + \boldcbar_2 + \boldgbar_2 
    = \boldtau - \boldD_{21} \boldsigma
\end{equation}
where $\boldcbar_2 = \boldc_2 - \boldD_{21}\boldc_1$, $\boldgbar_2 = \boldg_2 - \boldD_{21}\boldg_1$ and 
\begin{equation}
    \boldDbar_{22} = \boldD_{22} - \boldD_{21}\boldD_{12}
    \label{bar-D22-definition}
\end{equation}
The matrix $\boldDbar_{22}$ is the positive definite Schur complement of $\boldD$. Since $\|\boldD\|_2$ is lower bounded, it follows that the inverse matrix $\bar\boldD_{22}^{-1}$ is bounded by $\|\bar\boldD_{22}^{-1}\|_2 \leq \bar B_{22,\mathrm{inv}}$ for some positive constant $\bar B_{22,\mathrm{inv}}>0$.

Partial feedback linearization is then achieved with the generalized force vector 
$\boldtau = \boldDbar_{22} \boldv + \boldcbar_2 + \boldgbar_2
$ where $\boldv$ is a transformed control vector. Insertion into \eqref{eq:M22bar_dynamics_QY} and then insertion of the result into \eqref{eq:dynamics_unactuated_Y} in combination with \eqref{g1-expression} gives the partially linearized system 
\begin{align}
    \boldyddot + \boldc_1  + \Omega_z^2(\boldy - \boldy_0)
    &=  - \boldE\boldsigma - \boldD_{12} \boldv \label{eq:CL_dynamics_unactuated}\\
    \boldyddot_0 &= \boldv - \boldDbar_{22}^{-1} \boldD_{21} \boldsigma
    \label{eq:CL_dynamics_actuated}
\end{align}
where $\boldsigma$ is an unknown generalized disturbance force on the load and the matrix 
\begin{equation}\label{E-definition}
    \boldE(\boldp) = \boldD_{12}(\boldp)\boldDbar_{22}(\boldp)^{-1}\boldD_{21}(\boldp) - \eye
\end{equation} 
is bounded by $\|\boldE\|_2\leq B_E$ for some $B_E>0$ since the operator norms of $\boldD_{12}$, $\boldDbar_{22}^{-1}$ and $\boldD_{21}$ are bounded.

Let the desired crane tip position be $\boldy_{0d}$ and let the control deviation be $\boldytilde_0 = \boldy_0 - \boldy_{0d}$. The transformed control vector for the actuated crane tip is set to
\begin{equation}
\label{y0-control-v}
    \boldv = \boldyddot_{0d} - k_{d0}\boldytildedot_0 - k_{p0} \boldytilde_0
\end{equation}
where $k_{d0}>0$ and $k_{p0}>0$ are feedback gains. This gives
$ \Ddot{\boldytilde}_0 + k_{d0} \boldytildedot_0 + k_{p0} \boldytilde_0 = - \boldDbar_{22}^{-1} \boldD_{21} \boldsigma
$, which is an exponentially stable system when $\boldsigma=\zero$. 

Partial feedback linearization was originally formulated in \cite{Spong1994} so that the unactuated part was controlled with the desired acceleration $\ddot\boldy_{d0}$ of the actuated part. Here, this means that the dynamics of $\boldy$ as given by \eqref{eq:CL_dynamics_unactuated} would be controlled with the $\boldv$ vector. This was used in crane control in \cite{Tysse2022}. 
In this paper, we propose a modified version of partial feedback linearization where the unactuated part is controlled with the desired position $\boldy_{d0}$ of the crane tip. This leads to improved tracking performance for the load and allows for the use of nonparametric adaptive control. We start the development by rewriting equation \eqref{eq:CL_dynamics_unactuated} in the form
\begin{align}
    \boldyddot +\Omega_z^2\boldy
    &= \Omega_z^2\boldy_{0d} + \Omega_z^2\boldytilde_0
    - \boldE\boldsigma - \boldD_{12} \boldv - \boldc_1
    \nonumber\\
    &= \Omega_z^2\boldy_{0d} - \boldh
    \label{nominal-dynamics-unactuated}
\end{align} 
and use $\Omega_z^2\yv_{0d}$ is the control variable. Here $\boldh$ is a vector of unknown disturbance terms given by
\begin{equation}\label{h-definition}
    \boldh = - \Omega_z^2\boldytilde_0 + \boldE\boldsigma + \boldD_{12} \boldv + \boldc_1
\end{equation}
The system \eqref{nominal-dynamics-unactuated} is controlled by setting the control variable $\Omega_z^2\yv_{0d}$ to
\begin{equation}
    \Omega_z^2\yv_{0d} 
    = -k_p\yt-k_d\dot\yt + \ddot\yv_d +\boldu + \Omega_z^2\yv
    \label{y0d-constroller}
\end{equation}
where the control deviation is denoted $\boldytilde = \boldy - \boldy_d$ and $\boldy_d$ is the desired load mass position, while $k_d$ and $k_p$ are positive feedback gains and $\boldu$ is the nonparametric adaptive compensation. 
Insertion of \eqref{y0d-constroller} into \eqref{nominal-dynamics-unactuated} and insertion of \eqref{y0-control-v} into \eqref{eq:CL_dynamics_actuated} give the closed-loop dynamics of the partially feedback linearized system as
\begin{align}
    \ddot\boldytilde + k_d\boldytildedot + k_p \boldytilde 
    &=  \boldu - \boldh
    \label{eq:system_closed_loop_y}\\
    \Ddot{\boldytilde}_0 + k_{d0} \boldytildedot_0 + k_{p0} \boldytilde_0 
    &= - \boldDbar_{22}^{-1} \boldD_{21} \boldsigma
    \label{eq:system_closed_loop_y0}
\end{align}
The disturbance term $\boldh$ in \eqref{h-definition} is written in the form 
$
    \boldh = \boldE\boldsigma - \boldeta - \boldzeta
$
where $\boldE$ is defined in \eqref{E-definition}, 
$
    \boldeta = \Omega_z^2\boldytilde_0 
    - \boldD_{12}(k_{d0} \boldytildedot_0 + k_{p0} \boldytilde_0)
    $ and $
    \boldzeta = \boldD_{12}\boldyddot_{0d} - \boldc_1
$. 
The perturbation terms $\boldeta$ and $\boldzeta$ are bounded by 
\begin{align}
    \|\boldeta\|_2 &\leq C_{\eta p}\|\boldytilde_0\|_2 + C_{\eta d}\|\boldytildedot_0\|_2
    \label{Lipschitz-bound-on-eta}
    \\
    \|\boldzeta\|_2 &\leq B_{12}\|\boldyddot_{0d}\|_2
    + C_{c1}\left(\|\dot\yv\|_2^2 + \|\dot\yv_0\|_2^2\right)
    \label{Lipschitz-bound-on-zeta}
\end{align}
where $C_{\eta p} = \omega_0^2 + k_{p0}B_{12}$ and $C_{\eta d} = k_{d0}B_{12}$ are positive constants. The following assumption is reasonable in view of \eqref{Lipschitz-bound-on-zeta}. 
\begin{assumption}
\label{assumption-bounded-zeta}
    The term $\boldzeta$ satisfies $\|\boldzeta(t)\|_2 \leq B_\zeta$ for all $t\geq 0$ for some $B_\zeta>0$, and $\|\boldE\boldsigma\|_2 \leq B_\sigma$ for some $B_\sigma>0$. 
\end{assumption}

\subsection{Tracking controller}\label{sec:tracking_controller}

We now propose a tracking controller without adaption.

\begin{proposition}
    Consider the system (\ref{eq:system_closed_loop_y}, \ref{eq:system_closed_loop_y0}) with $\boldu = \zero$ under Assumption~\ref{assumption-bounded-zeta}, which gives
\begin{align}
\label{eq:y_system_closed_loop_perturbation_1}
     \ddot\boldytilde + k_d\boldytildedot + k_p\boldytilde 
     &= \boldeta + \boldzeta- \boldE\boldsigma\\
    \ddot{\boldytilde}_0 + k_{d0} \boldytildedot_0 + k_{p0} \boldytilde_0 &= - \boldDbar_{22}^{-1} \boldD_{21} \boldsigma
\label{eq:y_system_closed_loop_perturbation_2}
\end{align}
The subsystem \eqref{eq:y_system_closed_loop_perturbation_2} is exponentially stable when $\boldsigma=\zero$, while (\ref{eq:y_system_closed_loop_perturbation_1}, \ref{eq:y_system_closed_loop_perturbation_2}) with state  $\boldz = [\boldx^\T,\boldx_0^\T]^\T$ where $\boldx = [\boldytilde^\T,\boldytildedot^\T]^\T$ and $\boldx_0 = [\boldytilde_0^\T,\boldytildedot_0^\T]^\T$ is uniformly ultimately bounded with a bound that is proportional to the bounded norm of the vector $[\boldzeta-\boldE\boldsigma,- \boldDbar_{22}^{-1} \boldD_{21} \boldsigma]^\T$.
\end{proposition}
{\it Proof:}
The subsystem \eqref{eq:y_system_closed_loop_perturbation_2} is obviously exponentially stable when $\boldsigma=\zero$. The system (\ref{eq:y_system_closed_loop_perturbation_1}, \ref{eq:y_system_closed_loop_perturbation_2}) is a perturbation of the system 
\begin{align}
     \ddot\boldytilde + k_d\boldytildedot + k_p\boldytilde 
     &= \boldeta    \label{eq:y_system_closed_loop_tracking-controller}\\
    \ddot{\boldytilde}_0 + k_{d0} \boldytildedot_0 + k_{p0} \boldytilde_0 &= \zero
\label{eq:y0_system_closed_loop_tracking-controller}
\end{align}
which is exponentially stable according to \cite[p.~537]{Khalil2002} since \eqref{eq:y0_system_closed_loop_tracking-controller} is exponentially stable and $\boldeta$ is Lipschitz in $[\boldytilde_0^\T,\boldytildedot_0^\T]^\T$, which is seen from \eqref{Lipschitz-bound-on-eta}. Since the system (\ref{eq:y_system_closed_loop_perturbation_1}, \ref{eq:y_system_closed_loop_perturbation_2}) is equal to the system (\ref{eq:y_system_closed_loop_tracking-controller}, \ref{eq:y0_system_closed_loop_tracking-controller}) plus a bounded nonvanishing perturbation $[\boldzeta-\boldE\boldsigma,- \boldDbar_{22}^{-1} \boldD_{21} \boldsigma]^\T$, it follows from \cite[Lemma 9.2]{Khalil2002} that (\ref{eq:y_system_closed_loop_perturbation_1}, \ref{eq:y_system_closed_loop_perturbation_2}) uniformly ultimately bounded with a bound that is proportional to the norm of the perturbation.  
\hboks




\subsection{Adaptive control}\label{sec:nonparametric_controller}

The nonparametric adaptive control law of \cite{Boffi2022} is applied to the crane control problem in this section. The combined crane and payload dynamics are given by (\ref{eq:system_closed_loop_y}, \ref{eq:system_closed_loop_y0}). The adaptive controller is applied to the load dynamics \eqref{eq:system_closed_loop_y}. Due to the partial feedback linearization, the crane dynamics \eqref{eq:system_closed_loop_y0} will not be influenced by the payload dynamics \eqref{eq:system_closed_loop_y}. This means that the crane dynamics \eqref{eq:system_closed_loop_y0} will have no impact on the stability of the adaptive controller, but will only contribute through the bounded disturbances $\boldE\boldsigma$, $\boldeta$ and $\boldzeta$. 

In the terminology of \cite{Boffi2022} the system dynamics is given by the closed-loop load dynamics \eqref{eq:system_closed_loop_y}. Let the state vector be $\boldx =[\boldx_1^\tr, \boldx_2^\tr]^\tr$ where $\boldx_1 = \boldy$ and $\boldx_2=\dot\boldy$. Let the desired state be $\boldx_d =[\boldx_{1d}^\tr, \boldx_{2d}^\tr]^\tr$, and let the error vector be $\bolde = \boldx-\boldx_d$, which is written $\bolde =[\bolde_1^\tr, \bolde_2^\tr]^\tr$ where $\bolde_1 = \yt$ and $\bolde_2=\dot\yt$. 
The system dynamics in state space formulation is 
\begin{equation}
    \dot\boldx = \f(\boldx,t) 
       + \B\left(\boldu - \boldh \right)
       \label{dynamics-ss}
\end{equation}
where $\boldB = [\zero,\eye]^\tr$ and
\begin{equation}
  \f(\boldx,t) 
        =   \bb \boldx_2 \\
           -k_p \bolde_1 -k_d\bolde_2 + \dot\boldx_{2d}(t)
            \eb
\end{equation}
It is seen that $\|\boldB\|_2=1$ and that $\f(\boldx)$ is locally bounded and Lipschitz in $\boldx$ uniformly in $t$. 

The nonparametric adaptive control law (\ref{adaptive-nonparametric-h-hat}, \ref{adaptive-nonparametric-update-beta-hat}) is used where
$\boldu = \hat\boldh = \boldPsi(\boldx)^{\T} \alphavh$ is used to compensate for the unknown disturbance $\boldh = \boldPsi(\boldx)^{\T} \alphav + \epsv$ as given by \eqref{h=Psi-alpha+epsilon}. The estimation error is then given by \eqref{nonparametric-estimation error} as $\hat\boldh(\boldx) - \boldh(\boldx) = \boldPsi(\boldx)^{\T} \alphavt - \epsv(\boldx)$. The resulting error system is
\begin{align}
    \dot\bolde &= \f_e(\bolde) 
       + \B\left( \boldPsi(\boldx)^{\T} \alphavt - \epsv \right)
       \label{error-dynamics-ss}
       \\
    \dot\alphavh &= -\gamma \boldPsi(\boldx) \B^\T \nabla Q(\bolde)
    \label{adaptive-law-ss}
\end{align}
where 
\begin{equation}
  \f_e(\bolde) 
        =   \bb \bolde_2 \\
           - k_p \bolde_1 - k_d\bolde_2 
            \eb
            \label{f-e-def}
\end{equation}
is locally bounded and Lipschitz in $\bolde$. 
It is noted that $\boldB^\tr\nabla Q(\bolde) = c\bolde_1 + \bolde_2$, which gives
\begin{equation}
    \|\boldB^\tr\nabla Q(\bolde)\|\leq k_g\|\bolde\|_2, \ k_g = \min(c,1)
    \label{B-nabla-Q-bound}
\end{equation}

\begin{proposition}
    \label{prop-unperturbed-error-Lyapunov}
    The nominal error dynamics 
    \begin{equation}
        \dot\bolde = \f_e(\bolde)
        \label{unperturbed-error-dynamics}
    \end{equation}
    are exponentially stable and admit a Lyapunov function $Q(\bolde)$ so that  $\nabla Q(\bolde)$ is locally bounded in $\bolde$ and locally Lipschitz in $\bolde$, $\nabla Q(\bolde)^\T\f_e(\bolde) \leq -k_Q\|\bolde\|_2^2$ and $k_1\|\bolde\|_2^2\leq Q(\bolde)\leq k_2\|\bolde\|_2^2$ where $k_Q$, $k_1$ and $k_2$ are positive constants. 
\end{proposition}
{\it Proof:} Let    
\begin{equation}\label{eq:lyapunov_funQ}
    Q(\bolde) =  \frac{1}{2} \bolde^{\T} \boldP \bolde
\end{equation}
where the positive definite matrix $\boldP$ is given by 
\begin{equation}
    \boldP = \bb
       (k_p + k_d c)\I & c\I \\
            c\I & \I
        \eb
\end{equation}
where $k_p>0$, $k_d>c>0$, $\det(\boldP) = k_p+k_dc-c^2>0$ and $\boldP$ has eigenvalues $k_2>k_1>0$ \cite{Wen1988}. The time derivative of ${Q}$ along the trajectories of \eqref{unperturbed-error-dynamics}  is
\begin{equation}
    \Qdot(\bolde) = \nabla^\T Q(\bolde)\f_e(\bolde) 
    = -ck_p\bolde_1^\tr\bolde_1 - k_c\bolde_2^\tr\bolde_2 
\end{equation}
where $k_c=k_d-c>0$. Then $\nabla Q(\bolde)=\boldP \bolde$ is locally Lipschitz in $\bolde$ and locally bounded in $\bolde$, 
\begin{align}
    \nabla^\T Q(\bolde)\f_e(\bolde) &\leq - k_Q\|\bolde\|_2^2, \ k_Q=\min\{ck_p,k_c\}
    \label{Q-bounds-1}\\
    k_1\|\bolde\|_2^2 &\leq  Q(\bolde) \leq k_2 \|\bolde\|_2^2
    \label{Q-bounds-2}
\end{align}
and exponentially stability of \eqref{unperturbed-error-dynamics} follows.
\hboks

It is noted that the system dynamics \eqref{dynamics-ss} and the error system \eqref{error-dynamics-ss} satisfies \cite[Assumptions 3.3, 3.4 and 3.7]{Boffi2022}, which follows from \eqref{dynamics-ss}, \eqref{error-dynamics-ss} and Proposition~\ref{prop-unperturbed-error-Lyapunov}.
\begin{theorem}
    \label{thm:ultimate-bound-b} 
    Consider the system given by \eqref{dynamics-ss}-\eqref{f-e-def} under Assumptions \ref{ass-Gaussian-kernel} and \ref{ass:x-and-alpha-bounds}. Fix $\delta\in (0,1)$, $B_h > 0$ and a positive integer $d$. 
    Then with probability at least $1-\delta$, $\|\bolde(t)\|_2$ is uniformly ultimately bounded, and $\limsup_{t\rightarrow\infty}\|\bolde(t)\|_2\leq \varepsilon$ for some $\varepsilon>0$ whenever
\begin{equation}\label{bound-on-d}
    d \geq \left(\frac{k_g}{\theta k_Q}\sqrt{\frac{k_2}{k_1}}
    \frac{4B_h}{\varepsilon}\left(\frac{B_X\sqrt{n}}{\sigma} + \sqrt{m} + g(\delta) \right)\right)^2
\end{equation}
\vspace{-3mm}
\end{theorem}  
{\it Proof:}
Consider the nonnegative function
\begin{equation}
    V = Q(\e) + \frac{1}{2\gamma}\alphavt^\T\alphavt
\end{equation}
The time derivative of $V$ along the trajectories of (\ref{error-dynamics-ss}, \ref{adaptive-law-ss}) is
\begin{align}
    \dot V 
    &= \nabla^\T Q(\bolde) \left(\f_e(\bolde) + \B\left( \boldPsi(\boldx)^{\T} \alphavt - \epsv \right)\right) \nonumber\\
    &\quad\quad  -\alphavt^\tr\boldPsi(\boldx) \B^\T \nabla Q(\bolde) 
    \nonumber\\
    &= \nabla^\T Q(\bolde) \f_e(\bolde) 
    - \nabla^\T Q(\bolde)\B \epsv \nonumber\\
    &\leq -k_Q\|\bolde\|_2^2 + k_gB_\epsilon\|\bolde\|_2
    \nonumber\\
    &= -k_Q(1-\theta)\|\bolde\|_2^2 - k_Q\theta \|\bolde\|_2^2 
    + k_gB_\epsilon\|\bolde\|_2
    \nonumber\\
    &\leq -k_Q(1-\theta)\|\bolde\|_2^2,
    \quad \forall \|\bolde\|_2 \geq k_Q^{-1}k_gB_\epsilon/\theta
    \label{dot-V-no-deadzone}
\end{align}
where $0<\theta<1$. The first inequality follows from \eqref{epsilon-bound}, \eqref{B-nabla-Q-bound} and \eqref{Q-bounds-1} and Schwarz inequality. It follows from \cite[Lemma 9.2]{Khalil2002} that $\|\bolde(t)\|_2$ is uniformly ultimately bounded since for some finite $T$
\begin{align}
    \|\bolde(t)\|_2 &\leq k e^{-\gamma (t-t_0)}\|\bolde(t_0)\|,\quad t \leq T
     \\
    \|\bolde(t)\|_2 &\leq b,\quad t \geq T
    \label{ultimate-bound}
\end{align}
where $k = \sqrt{k_2/k_1}$, $\gamma = (1-\theta)k_Q/(2k_2)$ and 
$
    b = \frac{k_g}{\theta k_Q}\sqrt{\frac{k_2}{k_1}} B_\epsilon
$
where $B_\epsilon$ is given by \eqref{B-epsilon-def}. It follows that $\limsup_{t\rightarrow\infty}\|\bolde(t)\|_2\leq \varepsilon$ whenever \eqref{bound-on-d} is satisfied. 
\hboks

\subsection{Adaption with deadzone and saturation}

The approximation of the unknown disturbance $\boldh$ will have a nonzero approximation error, and therefore it makes sense to use a deadzone function in the parameter update and combine this with saturation to limit the effect of noise \cite{Boffi2022}. We used the following piecewise linear deadzone and saturation function:
\begin{equation}
\label{def-F}
       F(x) = 
    \begin{cases}
		0, & x\leq \Delta\\
            \frac{x-\Delta}{2\mu}, & \Delta < x < \Delta+2\mu\\
            1, & \Delta+2\mu \leq x
    \end{cases} 
\end{equation}
for positive constants $\Delta$ and $\mu$. This function is continuous and locally Lipschitz, and $F(x) = (d/dx)G(x)$ where
\begin{equation}
       G(x) = 
    \begin{cases}
		0, & x\leq \Delta\\
            \frac{(x-\Delta)^2}{4\mu}, & \Delta < x < \Delta+2\mu\\
            x - (\Delta+\mu), & \Delta+2\mu \leq x
    \end{cases} 
\end{equation}
The adaption law with saturation is set to
\begin{equation}
\label{adaptive-law-ss-saturation}
    \dot\alphavh = -\gamma F(Q(\bolde)) \boldPsi(\boldx) \B^\T \nabla Q(\bolde)
\end{equation}
which is equal to the update law \eqref{adaptive-law-ss} multiplied with the deadzone and saturation function $F(Q(\bolde))$.

\begin{theorem}
    Consider the system given by \eqref{dynamics-ss}-\eqref{error-dynamics-ss}, \eqref{f-e-def} and \eqref{adaptive-law-ss-saturation} under Assumptions \ref{ass-Gaussian-kernel} and \ref{ass:x-and-alpha-bounds}. Fix $\delta\in (0,1)$, $B_h > 0$, $\Delta>0$ and a positive integer $d$. 
    Then with probability at least $1-\delta$, $\|\bolde(t)\|_2$ is uniformly ultimately bounded, and $\lim_{t\rightarrow\infty}\sup\|\bolde(t)\|_2\leq \sqrt{\Delta/k_1} $ whenever
\begin{equation}\label{bound-on-delta}
    \Delta \geq k_2(2k_Q^{-1}k_g B_\epsilon)^2
\end{equation}
\end{theorem}  
{\it Proof:} The proof is based on the proof of \cite[Theorem 6.4]{Boffi2022}. Consider the nonnegative function
\begin{equation}
    V = G(Q(\bolde)) + \frac{1}{2\gamma}\alphavt^\tr\alphavt
\end{equation}
Let $\Delta \geq k_2(2k_Q^{-1}k_g B_\epsilon)^2$. Then the time derivative of $V$ along the trajectories of \eqref{error-dynamics-ss} and \eqref{adaptive-law-ss-saturation} is
\begin{align}
    \dot V 
    &= F(Q(\bolde))
    \left( \nabla^\T Q(\bolde) \f_e(\bolde) - \nabla^\T Q(\bolde) \B_e\epsv\right) \nonumber \\
    &\leq  - F(Q(\bolde))\left( k_Q\|\bolde\|_2^2 
       - k_g B_\epsilon\|\bolde\|_2\right)  
    \nonumber \\
    &=  - F(Q(\bolde))k_Q \left( \|\bolde\|_2 
       - k_Q^{-1}k_g B_\epsilon\right) \|\bolde\|_2
    \nonumber \\
    &\leq  - F(Q(\bolde)) k_Q\left( \frach(k_2^{-1} \Delta)^{1/2} 
       \right)(k_2^{-1} \Delta)^{1/2} \nonumber\\
    &\leq  - \frach F(Q(\bolde)) k_Q k_2^{-1} \Delta
    \label{dot-V-saturation}
\end{align}
The first equality and the first inequality follows from \eqref{dot-V-no-deadzone}. The second inequality follows since $F(Q(\bolde))>0$ implies $Q(\bolde)>\Delta$ and $\|\bolde\|_2 \geq (k_2^{-1} Q)^{1/2} > (k_2^{-1} \Delta)^{1/2}$ where \eqref{Q-bounds-2} is used, and $k_Q^{-1}k_g B_\epsilon \leq \frach(k_2^{-1} \Delta)^{1/2}$ by assumption. Integration of \eqref{dot-V-saturation} gives
$
    \int_0^\infty F(Q(\bolde))dt \leq \frac{2V(0)}{k_Qk_2^{-1} \Delta}
$
where it is used that $V(0) - V(t)\leq V(0)$ for all $t\geq 0$. $\bolde(t)$ is uniformly continuous since $\dot\bolde$ given by \eqref{f-e-def} is bounded for exponentially stable $\bolde$. Since $Q(\cdot)$ is locally Lipschitz this implies that $Q(\bolde)$ is uniformly continuous, and since $F(\cdot)$ is locally Lipschitz, it follows that $F(Q(\bolde))$ is uniformly continuous.  Since $ \int_0^\infty F(Q(\bolde))dt <\infty$ and $F(Q(\bolde))$ is uniformly continuous, it follows from Barbalat's lemma that $\lim_{t\rightarrow\infty}F(Q(\bolde)) = 0$. It follows from \eqref{def-F} that $\limsup_{t\rightarrow\infty}Q(\bolde) \leq \Delta$ and from \eqref{Q-bounds-2} that 
$
\limsup_{t\rightarrow\infty}\|\bolde(t)\|_2 \leq \sqrt{\Delta/k_1} 
$
\hboks

{\it Remark:} The bound on $\Delta$ in \eqref{bound-on-delta} is different from the one used in \cite[Theorem 6.4]{Boffi2022}. It is noted that if $\Delta$ is set to the smallest allowable value $\Delta = k_2(2k_Q^{-1}k_g B_\epsilon)^2$ of \eqref{bound-on-delta}, then
$
    \limsup_{t\rightarrow\infty}\|\bolde(t)\|_2\leq 2\frac{k_g}{k_Q}\sqrt{\frac{k_2}{k_1}} B_\epsilon
$
which is equal to the ultimate bound in Theorem~\eqref{thm:ultimate-bound-b} when $\theta=1/2$.

\section{Experiments}\label{sec:experiments}

The proposed tracking controller and the nonparametric adaptive controller were evaluated in both simulation and experiments. The simulation studies were implemented in Simulink, and the experiments were performed using a KUKA KR120 industrial robot in place of a crane, where the end effector of the robot was used as the suspension point of the payload. A model of the plant is is presented in Figure~\ref{fig:figure_crane_experiments}, and the parameters of the spherical pendulum with a moving suspension point for both the simulation studies and the experiments are presented in Table \ref{tab:system_params}.

\begin{figure}[h]
    \vspace{1mm}
    \centering
    \input{figures/models/figure_crane_experiments}
    \caption{Model of the crane system showing the main components of the test setup and notation used to present the results.}
    \label{fig:figure_crane_experiments}
\end{figure}

\begin{table}[h]
\vspace{1mm}
\caption{Physical system parameters}
\centering
\label{tab:system_params}
\begin{tabular}{@{\extracolsep\fill}m{117pt}ccc}
\toprule
Parameter & Symbol & Value & Unit\\
\midrule
Payload mass & $m$ & $4.0$ & \si{\kilogram}\\
Cable length & $L$ & $1.255$ & \si{\metre}\\
Gravitational acc. & $g$ & $9.81$ & \si{\metre\per\second\squared}\\
Natural frequency & $\omega_0$ & $2.796$ & \si{\radian\per\second\squared}\\
\bottomrule
\end{tabular}
\end{table}

The proposed Cartesian tracking controller was tuned as a damped harmonic oscillator by selecting the undamped natural frequency ${\omega_c}$ and the relative damping ${\zeta_c}$. This was used to determine the controller gains as ${k_{p,c} = \omega_c^2}$ and ${k_{d,c} = 2 \zeta_c \omega_c}$. The parameters of the proposed Cartesian tracking controller are given in Table \ref{tab:tracking_params}.
\begin{table}[h]
\caption{Cartesian tracking controller parameters}
\centering
\label{tab:tracking_params}
\begin{tabular}{@{\extracolsep\fill}m{117pt}ccc}
\toprule
Parameter & Symbol & Value & Unit\\
\midrule
Undamped natural frequency & $\omega_c$ & $2.796$ & \si{\radian\per\second\squared}\\
Relative damping & $\zeta_c$ & $0.2$ & -\\
Proportional gain & $k_{p,c}$ & $7.817$ & -\\
Derivative gain & $k_{d,c}$ & $1.118$ & -\\
\bottomrule
\end{tabular}
\end{table}

\subsection{Comparison of angular and Cartesian formulation}

\begin{figure}[h]
    \begin{subfigure}[h]{0.48\columnwidth}
        \centering
        \includegraphics[width=\textwidth]{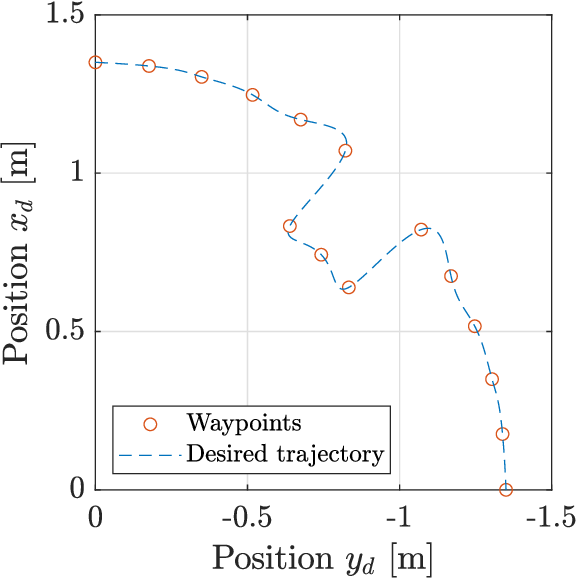}
        \caption{Reference trajectory ($xy$-plot)}
        \label{fig:ang_vs_cart_reference_position_scatter}
    \end{subfigure}
    \hfill
    \begin{subfigure}[h]{0.48\columnwidth}
        \centering
        \includegraphics[width=\textwidth]{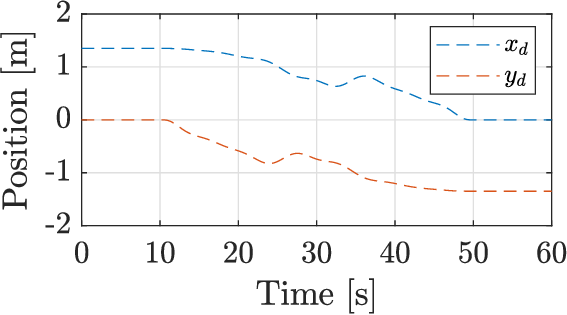}
        \includegraphics[width=\textwidth]{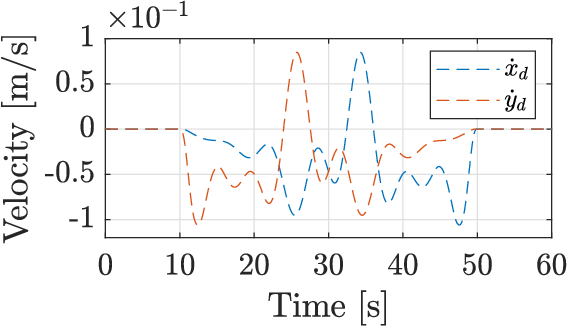}
        \includegraphics[width=\textwidth]{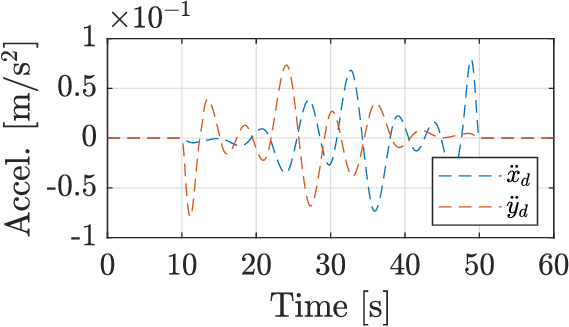}
        \caption{Reference trajectory time history}
        \label{fig:ang_vs_cart_reference_trajectory}
    \end{subfigure}
    \caption{Reference trajectory that was used in simulations for comparison of the Cartesian and angular controllers}
    \label{fig:ang_vs_cart_reference}
    \vspace{-2mm}
\end{figure}

A comparative study was performed where the proposed Cartesian tracking controller given by \eqref{y0d-constroller} where $\boldu = \zero$ was compared with the exponentially stabilizing damping controller presented in \cite{Tysse2022}. A reference trajectory was generated to simulate an obstacle avoidance scenario. The reference trajectory was a ${\ang{90}}$ rotation of crane about the vertical axis of the base frame. The resulting payload trajectory started with zero velocity at $ {x_0 = \SI{1.35}{\metre}, y_0 = \SI{0}{\metre}}$, and ended with zero velocity at ${x_T = \SI{0}{\metre}, y_T = \SI{-1.35}{\metre}}$. The duration of the trajectory was ${T = \SI{40}{\second}}$. A ${\SI{10}{\second}}$ buffer with zero velocity was added before the start and after the end of the reference trajectory. An obstacle was placed midway in the reference trajectory, and a set of waypoints was generated to avoid the obstacle by using a minimum jerk planner in MATLAB. An $xy$-plot of the reference trajectory is shown in Figure~\ref{fig:ang_vs_cart_reference_position_scatter}. The corresponding position, velocity, and acceleration profiles of the reference trajectory are shown in Figures~\ref{fig:ang_vs_cart_reference_trajectory}.

The angular controller used the exponentially stabilizing damping controller presented in \cite{Tysse2022} for the crane load combined with a tracking controller \cite{Tysse2021} for the suspension point. The angular damping controller was tuned according to \cite{Tysse2022} with the undamped natural frequency ${\omega_d^2 = k_{p,d} + \omega_0^2}$ and damping ratio ${\zeta_d = k_{d,d} / 2 \omega_d}$. The suspension point tracking controller was then tuned according to \cite{Tysse2021}, selecting ${\omega_t = \omega_d / 5}$ and ${\zeta_t \in [0.7,1]}$ to get the controller gains ${k_{p,t} = \omega_t^2}$ and ${k_{d,t} = 2 \zeta_t \omega_t}$. The controller parameters are given in Table \ref{tab:angular_damping_controller} and Table \ref{tab:angular_tracking_controller}, where the proportional gain for the damping controller was set to $k_{p,d} = 0$ for an undamped natural frequency ${\omega_d = \omega_0}$.

\begin{table}[h]
\caption{Angular damping controller parameters}
\centering
\label{tab:angular_damping_controller}
\begin{tabular}{@{\extracolsep\fill}m{117pt}ccc}
\toprule
Parameter & Symbol & Value & Unit\\
\midrule
Undamped natural frequency & $\omega_d$ & $2.796$ & \si{\radian\per\second\squared}\\
Relative damping & $\zeta_d$ & $0.2$ & -\\
Proportional gain & $k_{p,d}$ & $0$ & -\\
Derivative gain & $k_{d,d}$ & $1.118$ & -\\
\bottomrule
\end{tabular}
\end{table}

\begin{table}[h]
\vspace{1mm}
\caption{Suspension point tracking controller parameters}
\centering
\label{tab:angular_tracking_controller}
\begin{tabular}{@{\extracolsep\fill}m{117pt}ccc}
\toprule
Parameter & Symbol & Value & Unit\\
\midrule
Undamped natural frequency & $\omega_t$ & $0.559$ & \si{\radian\per\second\squared}\\
Relative damping & $\zeta_t$ & $1$ & -\\
Proportional gain & $k_{p,t}$ & $0.313$ & -\\
Derivative gain & $k_{d,t}$ & $1.118$ & -\\
\bottomrule
\end{tabular}
\vspace{2mm}
\end{table}

\begin{figure}[h]
    \vspace{2mm}
    \centering
    \begin{subfigure}[h]{0.48\columnwidth}
    \centering
        \includegraphics[width=\textwidth]{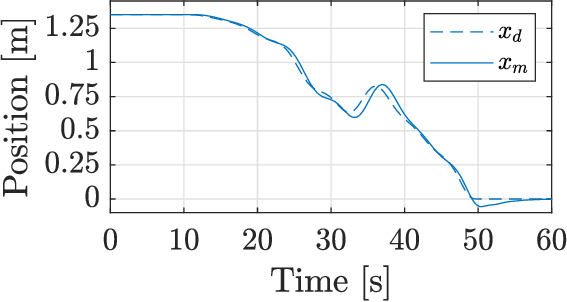}
    \end{subfigure}
    \hfill
    \begin{subfigure}[h]{0.48\columnwidth}
    \centering
        \includegraphics[width=\textwidth]{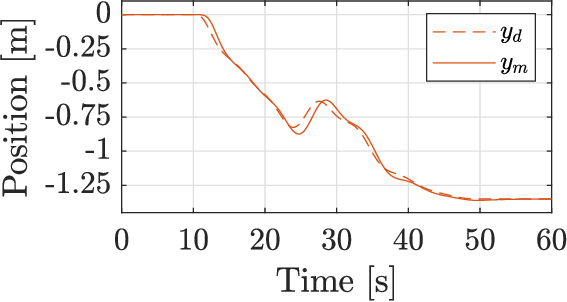}
    \end{subfigure}
    \caption{Angular controller tracking performance}
    \label{fig:angular_mass_position}
\end{figure}

\begin{figure}[h]
    \centering
    \begin{subfigure}[h]{0.48\columnwidth}
    \centering
        \includegraphics[width=\textwidth]{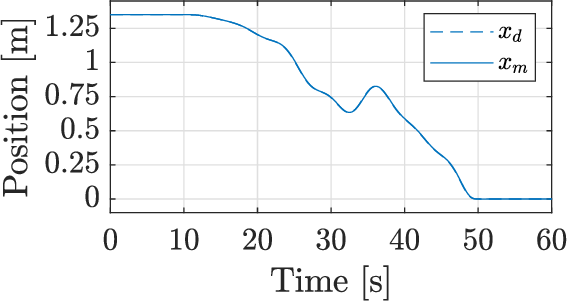}
    \end{subfigure}
    \hfill
    \begin{subfigure}[h]{0.48\columnwidth}
    \centering
        \includegraphics[width=\textwidth]{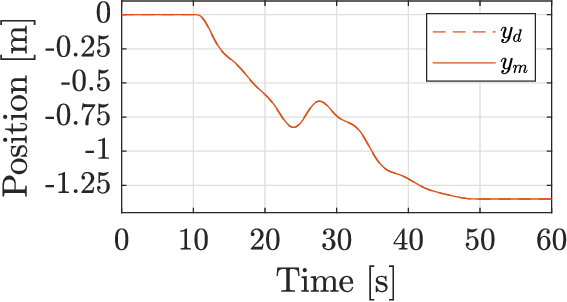}
    \end{subfigure}
    \caption{Cartesian controller tracking performance}
    \label{fig:cartesian_mass_position}
\end{figure}

The simulations demonstrated that tracking performance was significantly improved when the Cartesian controller was used compared to the angular controller. This was most evident during obstacle avoidance phase in the middle of the trajectory, where the angular controller gave significant overshoot, while the Cartesian controller tracked the trajectory accurately. The tracking performance is shown in Figures~\ref{fig:angular_mass_position} and \ref{fig:cartesian_mass_position}.


\begin{figure}[h]
    \centering
    \begin{subfigure}[h]{0.48\columnwidth}
    \centering
        \includegraphics[width=\textwidth]{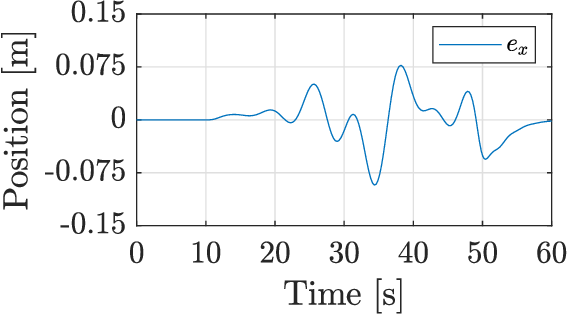}
    \end{subfigure}
    \hfill
    \begin{subfigure}[h]{0.48\columnwidth}
    \centering
        \includegraphics[width=\textwidth]{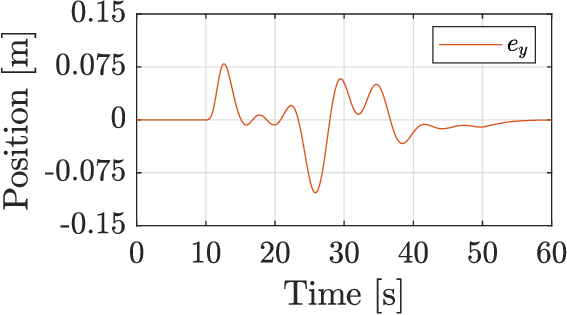}
    \end{subfigure}
    \caption{Position control error ${e_x}$ and ${e_y}$ for angular controller}
    \label{fig:angular_error_states}
\end{figure}

\begin{figure}[h]
    \centering
    \begin{subfigure}[h]{0.48\columnwidth}
    \centering
        \includegraphics[width=\textwidth]{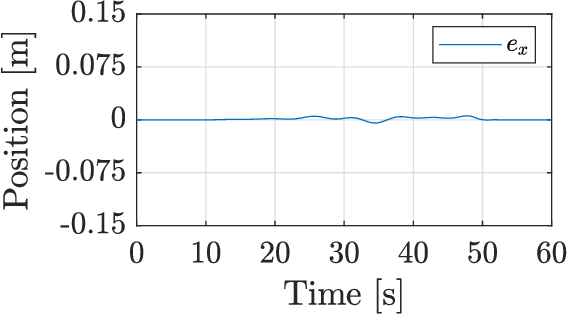}
    \end{subfigure}
    \hfill
    \begin{subfigure}[h]{0.48\columnwidth}
    \centering
        \includegraphics[width=\textwidth]{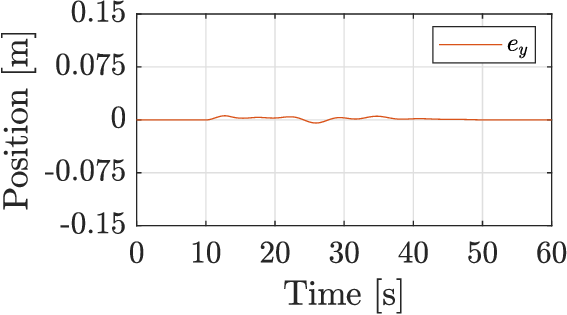}
    \end{subfigure}
    \caption{Position control error ${e_x}$ and ${e_y}$ for Cartesian controller}
    \label{fig:cartesian_error_states}
\end{figure}

The tracking error for the mass point was significantly smaller for the Cartesian controller than for the angular controller, which is seen from Figures~\ref{fig:angular_error_states} and \ref{fig:cartesian_error_states} and  Table~\ref{tab:ang_vs_cart_error_metrics}.

\begin{table}[h]
\caption{Tracking error metrics - Angular and Cartesian}
\centering
\label{tab:ang_vs_cart_error_metrics}
\begin{tabular}{@{\extracolsep\fill}m{44pt}ccc}
\toprule
Metric & Angular & Cartesian & Improvement [\%]\\
\midrule
MSE & $2.04 \cdot 10^{-3}$ & $1.35 \cdot 10^{-5}$ & $99.34$\\
MAE & $3.39 \cdot 10^{-2}$ & $3.20 \cdot 10^{-3}$ & $90.57$\\
\bottomrule
\end{tabular}
\end{table}


The improvement in tracking performance was not a consequence of a less efficient actuation of the suspension point. The comparison showed that the velocity and acceleration of the suspension point were comparable between the angular and Cartesian controllers. This is shown in Figures \ref{fig:angular_suspension_point} and \ref{fig:cartesian_suspension_point} below.

\begin{figure}[h]
    \centering
    \begin{subfigure}[h]{0.48\columnwidth}
    \centering
        \includegraphics[width=\textwidth]{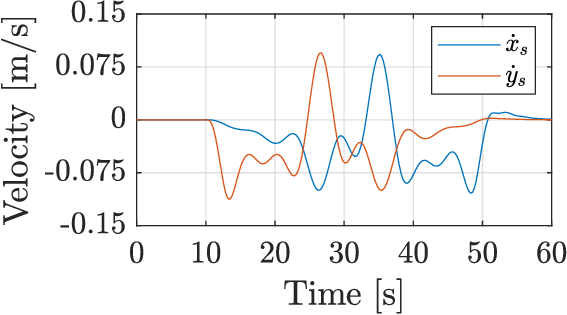}
    \end{subfigure}
    \hfill
    \begin{subfigure}[h]{0.48\columnwidth}
    \centering
        \includegraphics[width=\textwidth]{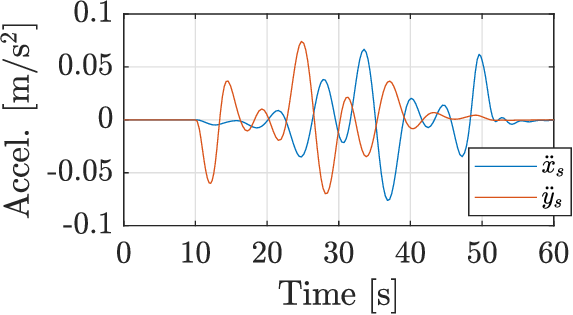}
    \end{subfigure}
    \caption{Suspension point velocity and acceleration for the angular controller}
    \label{fig:angular_suspension_point}
\end{figure}
\begin{figure}[h]
    \centering
    \begin{subfigure}[h]{0.48\columnwidth}
    \centering
        \includegraphics[width=\textwidth]{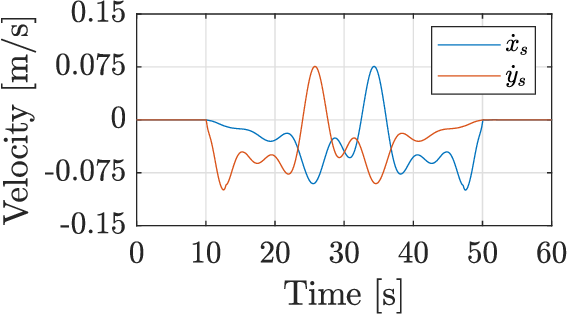}
    \end{subfigure}
    \hfill
    \begin{subfigure}[h]{0.48\columnwidth}
    \centering
        \includegraphics[width=\textwidth]{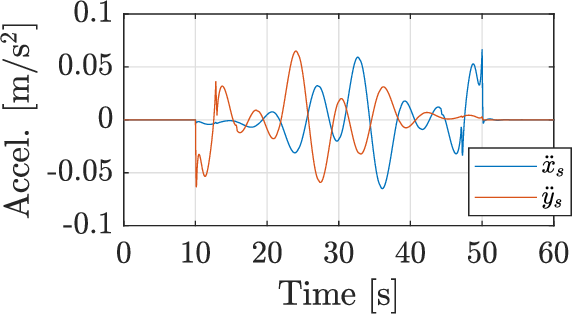}
    \end{subfigure}
    \caption{Suspension point velocity and acceleration for the Cartesian controller}
    \label{fig:cartesian_suspension_point}
\end{figure}

\subsection{Simulations and experiments with the nonparametric adaptive controller}

\begin{figure}[h]
    \begin{subfigure}[h]{0.48\columnwidth}
        \centering
        \includegraphics[width=\textwidth]{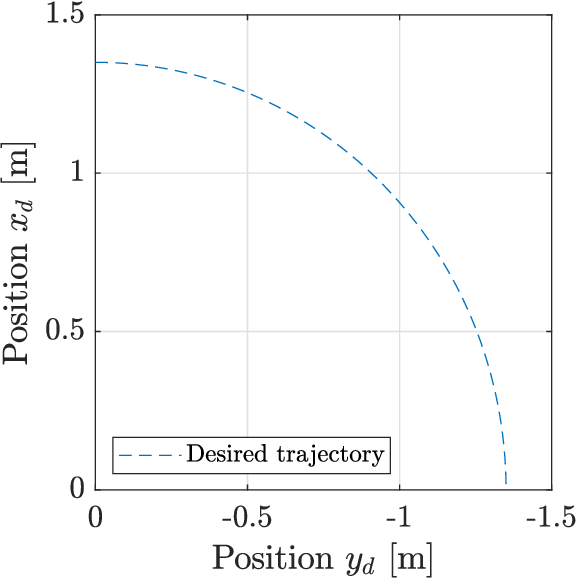}
        \caption{Reference trajectory ($xy$-plot)}
        \label{fig:reference_position_scatter}
    \end{subfigure}
    \hfill
    \begin{subfigure}[h]{0.48\columnwidth}
        \centering
        \includegraphics[width=\textwidth]{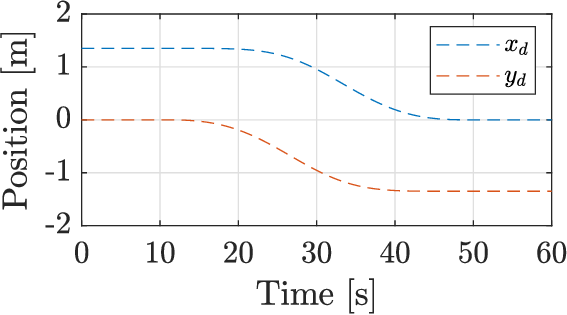}
        \includegraphics[width=\textwidth]{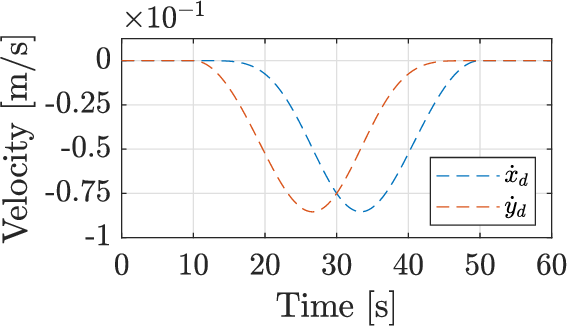}
        \includegraphics[width=\textwidth]{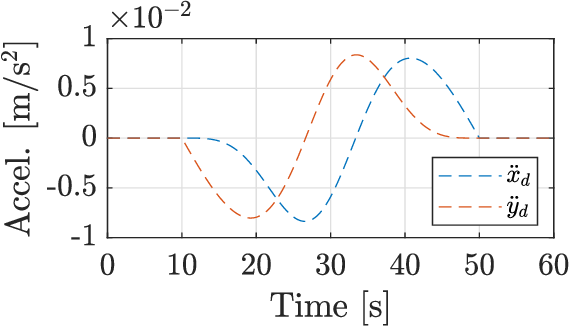}
        \caption{Reference trajectory time histories}
        \label{fig:reference_trajectory_profile}
    \end{subfigure}
    \caption{Reference trajectory for the adaptive control simulations and experiments}
    \label{fig:reference_trajectory}
\end{figure}

The nonparametric adaptive controller was compared to the non-adaptive Cartesian tracking controller in both simulations and experiments. The same reference trajectory of ${T = \SI{40}{\second}}$ duration was used as in the simulation study of the previous section, but in this case the time history was different, and there was no obstacle in the middle of the trajectory. The reference trajectory started with zero velocity at ${x_0 = \SI{1.35}{\metre}}$, ${y_0 = \SI{0}{\metre}}$, and ended with zero velocity at ${x_T = \SI{0}{\metre}}$, ${y_T = \SI{-1.35}{\metre}}$ as shown in the $xy$-plot of Figure~\ref{fig:reference_position_scatter}. A smooth sinusoidal acceleration profile was used to limit the jerk of the reference trajectory (Figure~\ref{fig:reference_trajectory_profile}). A ${\SI{10}{\second}}$ buffer with zero velocity was added before the start and after the end of the reference trajectory. 

The main disturbance to be compensated for by the adaptive controller was due to a sinusoidal motion of the base of the crane, which was similar to the wave-induced motion of a crane base on a ship deck. This sinusoidal motion was along the $y$-axis of the world frame $n$, with a frequency equal to the natural frequency of the pendulum ${\omega_0 = \SI{2.796}{\radian\per\second\squared}}$ and an amplitude of ${a = \SI{0.5}{\metre}}$. 

\subsubsection{Simulation study}

The crane with the nonparametric adaptive controller was simulated in Simulink. The parameters of the nonparametric adaptive controller used in the simulation are given in Table \ref{tab:adaptive_params_sim}.
\begin{table}[h]
\caption{Nonparametric adaptive controller parameters - Simulation}
\centering
\label{tab:adaptive_params_sim}
\begin{tabular}{@{\extracolsep\fill}m{127pt}ccc}
\toprule
Parameter & Symbol & Value & Unit\\
\midrule
Number of features & $d$ & $100$ & -\\
Kernel width & $\sigma$ & $1.5$ & -\\
Learning rate & $\gamma$ & $9$ & -\\
Lyapunov constant & $c$ & $0.5$ & -\\
\bottomrule
\end{tabular}
\vspace{-3mm}
\end{table}

\begin{figure}[h]
    \centering
    \begin{subfigure}[h]{0.48\columnwidth}
    \centering
        \includegraphics[width=\textwidth]{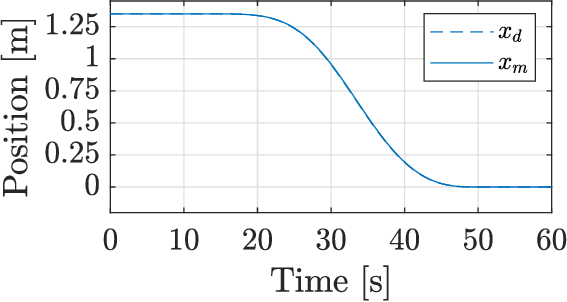}
    \end{subfigure}
    \hfill
    \begin{subfigure}[h]{0.48\columnwidth}
    \centering
        \includegraphics[width=\textwidth]{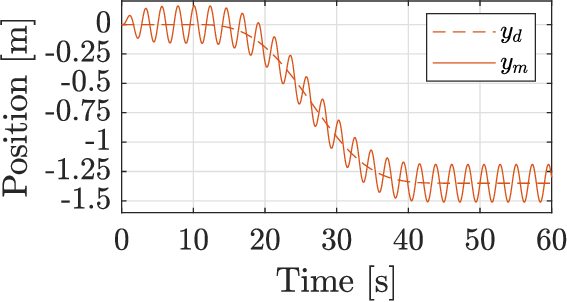}
    \end{subfigure}
    \caption{Simulation tracking results without adaption}
    \label{fig:sim_mass_position_without_learning}
\end{figure}

\begin{figure}[h]
    \centering
    \begin{subfigure}[h]{0.48\columnwidth}
    \centering
        \includegraphics[width=\textwidth]{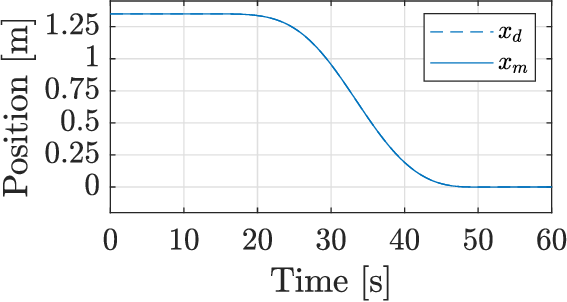}
    \end{subfigure}
    \hfill
    \begin{subfigure}[h]{0.48\columnwidth}
    \centering
        \includegraphics[width=\textwidth]{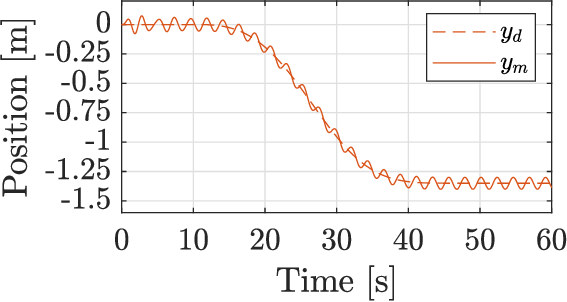}
    \end{subfigure}
    \caption{Simulation tracking results with adaption}
    \label{fig:sim_mass_position_with_learning}
\end{figure}

The simulation results showed that the nonparametric adaptive controller gave a significant improvement in tracking performance compared to the Cartesian tracking controller. The effect of the sinusoidal motion of the base was significantly reduced, which improved the tracking performance of the crane load in the $y$-direction. Furthermore, the nonparametric adaptive controller also improved the tracking accuracy in the $x$-direction. Figures~\ref{fig:sim_mass_position_without_learning} and \ref{fig:sim_mass_position_with_learning} show the simulated system without adaption and with adaption enabled, respectively.


\begin{figure}[h]
    \centering
    \begin{subfigure}[h]{0.48\columnwidth}
        \centering
        \includegraphics[width=\textwidth]{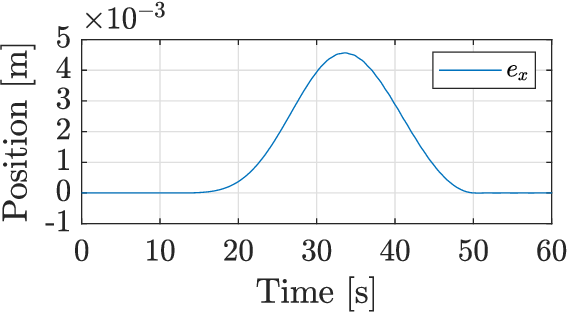}
    \end{subfigure}
    \hfill
    \begin{subfigure}[h]{0.48\columnwidth}
        \centering
        \vspace{3pt}
        \includegraphics[width=\textwidth]{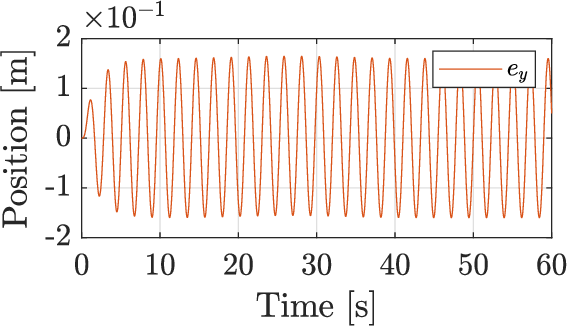}
    \end{subfigure}
    \caption{Simulation study error position without adaption}
    \label{fig:sim_error_without_learning}
\end{figure}

\begin{figure}[h]
    \centering
    \begin{subfigure}[h]{0.48\columnwidth}
        \centering
        \includegraphics[width=\textwidth]{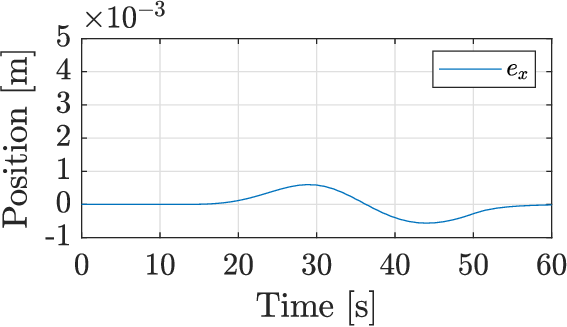}
    \end{subfigure}
    \hfill
    \begin{subfigure}[h]{0.48\columnwidth}
        \centering
        \vspace{3pt}
        \includegraphics[width=\textwidth]{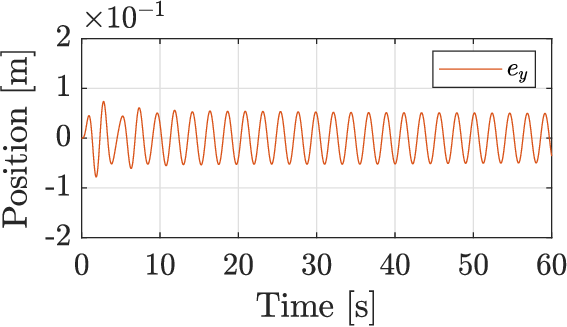}
    \end{subfigure}
    \caption{Simulation study error position with adaption}
    \label{fig:sim_error_with_learning}
\end{figure}

The reduction in position tracking error is illustrated in Figures~\ref{fig:sim_error_without_learning} and \ref{fig:sim_error_with_learning}, where the position error in the $x$- and $y$-directions are shown for the non-adaptive and adaptive case. The improvement is quantified in Table~\ref{tab:sim_tracking_error_metrics}.


\begin{table}[h]
\caption{Simulation - Tracking error metrics}
\centering
\label{tab:sim_tracking_error_metrics}
\begin{tabular}{@{\extracolsep\fill}m{38pt}ccc}
\toprule
Metric & W/o learn. & With learn. & Improvement [\%]\\
\midrule
MSE & $1.14 \cdot 10^{-2}$ & $1.50 \cdot 10^{-3}$ & $86.83$\\
MAE & $9.47 \cdot 10^{-2}$ & $3.52 \cdot 10^{-2}$ & $62.79$\\
\bottomrule
\end{tabular}
\end{table}

\begin{figure}[h]
    \centering
    \begin{subfigure}[h]{0.48\columnwidth}
    \centering
        \includegraphics[width=\textwidth]{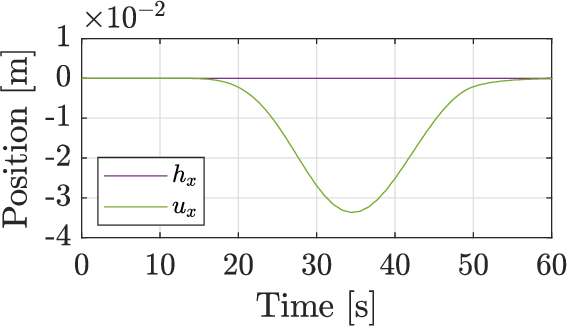}
    \end{subfigure}
    \hfill
    \begin{subfigure}[h]{0.48\columnwidth}
    \centering
        \includegraphics[width=\textwidth]{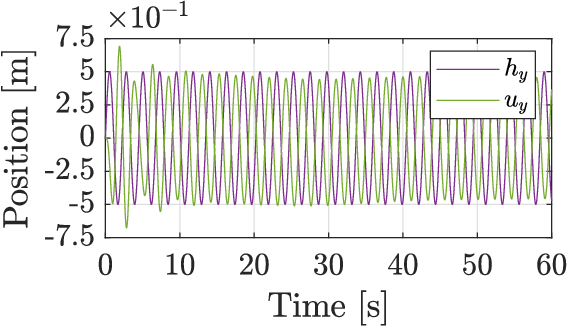}
    \end{subfigure}
    \caption{Simulation study disturbance and adaptive input with learning enabled}
    \label{fig:sim_disturb_vs_adaptive_in_with_learning}
\end{figure}

A closer inspection of the results further explains the improved tracking performance. The adaptive controller learns to counteract both the tracking error in the $x$-direction and the sinusoidal disturbance in the $y$-direction. This is shown in Figure \ref{fig:sim_disturb_vs_adaptive_in_with_learning}.

\subsubsection{Experimental validation}

The experiments were performed with a KUKA KR120 industrial robot which replaced the crane, using KUKA RobotSensorInterface to control the robot end effector (suspension point) in world frame coordinates and to read the position of the suspension point. For state feedback for the crane payload, a vision system using an Intel RealSense d435i camera was used with OpenCV to track the position of a ChArUco board attached to the crane payload. The position measurements of the payload were filtered using a low-pass filter, and the linear velocities of the payload were estimated using backward difference.
\begin{figure}[t]
    \centering
    \begin{tikzpicture}[scale=0.5,
    box/.style={line width=0.5pt},
    arrow/.style={line width=0.8pt},
    label/.style={font=\scriptsize}
]

\FPset\minBoxWidth{2.8}
\FPset\boxHeight{2}
\FPmul\halfBoxWidth{\minBoxWidth}{0.5}
\FPmul\halfBoxHeight{\boxHeight}{0.5}
\FPmul\quarterBoxWidth{\halfBoxWidth}{0.5}
\FPmul\quarterBoxHeight{\halfBoxHeight}{0.5}

\newcommand{\drawBox}[4]{
    \FPmax\boxWidth{\minBoxWidth}{#2}
    \FPmul\x{\boxWidth}{0.5}
    \FPmul\y{\boxHeight}{0.5}
    \FPsub\x{#3}{\x}
    \FPsub\y{#4}{\y}
    \fill[white] (\x,\y) rectangle ++(\boxWidth,\boxHeight);
    \draw[box] (\x,\y) rectangle ++(\boxWidth,\boxHeight);
    \node[label,align=center] at (#3,#4) {#1};
}


\FPset\borderWidth{16.8}
\FPset\borderHeight{9.5}
\FPset\margin{0.7}


\FPsub\upperX{\borderWidth}{\halfBoxWidth}
\FPsub\upperY{\borderHeight}{\halfBoxHeight}

\FPsub\upperX{\upperX}{\margin}
\FPsub\upperY{\upperY}{\margin}

\FPset\middleX{\upperX}
\FPmul\middleY{\borderHeight}{0.5}

\FPset\lowerX{\upperX}
\FPadd\lowerY{\halfBoxHeight}{\margin}


\fill[teal!10] (0,0) rectangle (\borderWidth,\borderHeight);

\FPsub\fastX{\middleX}{\halfBoxWidth}
\FPsub\fastY{\middleY}{\halfBoxHeight}
\FPsub\fastX{\fastX}{\margin}
\FPsub\fastY{\fastY}{\margin}

\fill[teal!50] (\fastX,\fastY) rectangle (\borderWidth,\borderHeight);


\drawBox{Crane}{0.0}{\upperX}{\upperY}

\drawBox{Payload}{0.0}{\middleX}{\middleY}

\FPset\trackingWidth{4}
\FPmul\trackingX{\borderWidth}{0.5}
\drawBox{Tracking\\controller}{\trackingWidth}{\trackingX}{\upperY}

\drawBox{Filter}{0.0}{\trackingX}{\lowerY}

\FPsub\adaptiveX{\trackingX}{\trackingWidth}
\FPset\adaptiveX{3.8}
\drawBox{Adaptive\\controller}{\trackingWidth}{\adaptiveX}{\middleY}


\newcommand{\drawLabel}[5]{
    \FPadd\posX{#2}{#4}
    \FPadd\posY{#3}{#5}
    \node[label,align=center] at (\posX,\posY) {#1};
}

\FPsub\y{\upperY}{\halfBoxHeight}
\FPadd\yy{\middleY}{\halfBoxHeight}
\draw[arrow,-latex] (\upperX,\y) -- (\middleX,\yy);

\FPadd\x{\trackingX}{\halfBoxWidth}
\FPsub\v{\middleY}{\halfBoxHeight}
\draw[arrow,dashed,-latex] (\middleX,\v) -- (\lowerX,\lowerY) -- (\x,\lowerY);

\drawLabel{Camera}{\lowerX}{\lowerY}{-1.1}{-0.35}
\drawLabel{$\boldy$}{\x}{\lowerY}{0.8}{0.3}

\FPmul\u{\trackingWidth}{0.5}
\FPadd\u{\trackingX}{\u}
\FPadd\v{\upperY}{\quarterBoxHeight}
\FPsub\uu{\upperX}{\halfBoxWidth}
\FPsub\vv{\upperY}{\quarterBoxHeight}
\FPmul\uuu{\u}{1.05}

\draw[arrow,-latex] (\u,\v) -- (\uu,\v);
\draw[arrow,latex-] (\u,\vv) -- (\uu,\vv);
\draw[arrow,-latex] (\uuu,\middleY) -- (\uu,\middleY);

\drawLabel{$\boldy_{0d}$}{\uu}{\v}{-0.85}{0.35}
\drawLabel{$\boldy_{0}$}{\u}{\vv}{0.8}{0.35}
\drawLabel{$\boldh$}{\uuu}{\middleY}{0.5}{0.35}

\FPmul\x{\trackingWidth}{0.5}
\FPsub\x{\trackingX}{\x}

\draw[arrow,-latex] (\margin,\v) -- (\x,\v);

\drawLabel{$\boldy_d,\boldydot_d,\boldyddot_d$}{\margin}{\v}{1.44}{0.4}

\FPsub\u{\trackingX}{\halfBoxWidth}

\draw[arrow,-latex] (\u,\lowerY) -- (\margin,\lowerY) -- (\margin,\vv) -- (\x,\vv);

\drawLabel{$\boldy,\boldydot$}{\u}{\lowerY}{-1}{0.35}

\FPset\circleDim{0.1}
\FPmul\x{\trackingWidth}{0.5}
\FPsub\x{\adaptiveX}{\x}

\filldraw[black] (\margin,\middleY) circle (\circleDim);
\draw[arrow,-latex] (\margin,\middleY) -- (\x,\middleY);


\FPset\jumpLength{1}
\FPmul\jumpLength{\jumpLength}{0.25}
\FPadd\b{\vv}{\jumpLength}
\FPsub\bb{\vv}{\jumpLength}

\filldraw[black] (\adaptiveX,\v) circle (\circleDim);

\draw[arrow,-latex] (\adaptiveX,\v) --  (\adaptiveX,\b) arc[start angle = 70,
                      end angle = -70,
                      x radius = \jumpLength cm,
                      y radius = \jumpLength cm] -- (\adaptiveX,\yy);

\drawLabel{$\boldy_d,\boldydot_d$}{\adaptiveX}{\yy}{1}{0.4}

\FPmul\x{\trackingWidth}{0.5}
\FPadd\x{\adaptiveX}{\x}

\draw[arrow,-latex] (\x,\middleY) -- (\trackingX,\middleY) -- (\trackingX,\y);

\drawLabel{$\boldu$}{\trackingX}{\y}{0.4}{-0.8}
    
\end{tikzpicture}
    \vspace{-2mm}
    \caption{Block diagram showing the test setup used for the experimental validation with the proposed control algorithm, camera and filter used for state feedback, and plant. Light color is the slow process, and dark color is the fast process.}
    \label{fig:figure_test_platform}
\end{figure}

The software was implemented in Python and was separated into a slow and fast process using multiprocessing. The slow process included the vision system, the tracking controller, and the nonparametric adaptive controller, and ran at ${\SI{30}{\hertz}}$, limited by the camera frame rate. The control input from the slow process was sent to the fast process running at ${\SI{250}{\hertz}}$ as required by the communication interface with the KR120 robot, sending position updates using KUKA RSI Ethernet. The test setup used for the experimental validation is illustrated in Figure~\ref{fig:figure_test_platform}.

\begin{table}[h]
\caption{Nonparametric adaptive controller parameters - Experimental validation}
\centering
\label{tab:adaptive_params_real}
\begin{tabular}{@{\extracolsep\fill}m{127pt}ccc}
\toprule
Parameter & Symbol & Value & Unit\\
\midrule
Number of features & $d$ & $1000$ & -\\
Kernel width & $\sigma$ & $0.5$ & -\\
Learning rate & $\gamma$ & $7$ & -\\
Lyapunov constant & $c$ & $0.5$ & -\\
Deadzone cutoff constant & $\Delta$ & $0.007$ & -\\
Deadzone smoothing constant & $\mu$ & $0.002$ & -\\
\bottomrule
\end{tabular}
\end{table}

Due to the noise level in the vision system, deadzones were implemented and a more conservative tuning of the nonparametric adaptive controller was used in the experiments. The parameters of the nonparametric adaptive controller used in the real experiment are given in Table \ref{tab:adaptive_params_real}.

\begin{figure}[h]
    \centering
    \begin{subfigure}[h]{0.48\columnwidth}
    \centering
        \includegraphics[width=\textwidth]{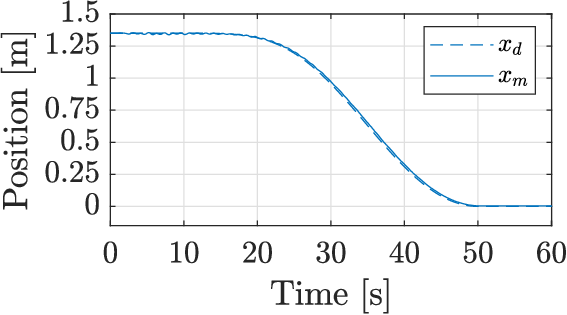}
    \end{subfigure}
    \hfill
    \begin{subfigure}[h]{0.48\columnwidth}
    \centering
        \includegraphics[width=\textwidth]{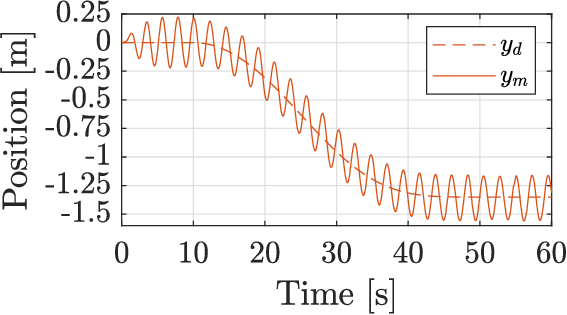}
    \end{subfigure}
    \caption{Experimental validation tracking results without learning}
    \label{fig:real_mass_position_without_learning}
\end{figure}

\begin{figure}[h]
    \centering
    \begin{subfigure}[h]{0.48\columnwidth}
    \centering
        \includegraphics[width=\textwidth]{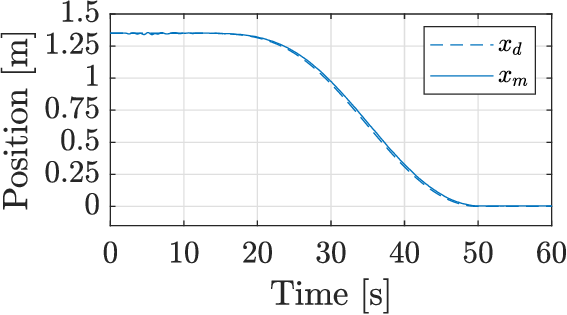}
    \end{subfigure}
    \hfill
    \begin{subfigure}[h]{0.48\columnwidth}
    \centering
        \includegraphics[width=\textwidth]{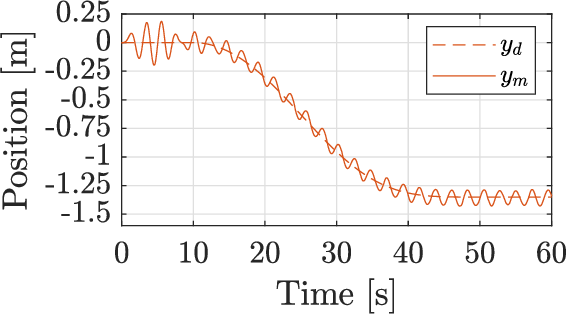}
    \end{subfigure}
    \caption{Experimental validation tracking results with learning}
    \label{fig:real_mass_position_with_learning}
\end{figure}

The experiments showed a significant improvement in tracking performance in the $y$-direction but a negligible improvement in the $x$-direction. The nonparametric adaptive controller was able to learn and cancel much of the disturbance, leading to a significant improvement in tracking performance. Figures~\ref{fig:real_mass_position_without_learning} and \ref{fig:real_mass_position_with_learning} show the system tracking performance without and with learning enabled, respectively.


\begin{figure}[h]
    \centering
    \begin{subfigure}[h]{0.48\columnwidth}
        \centering
        \includegraphics[width=\textwidth]{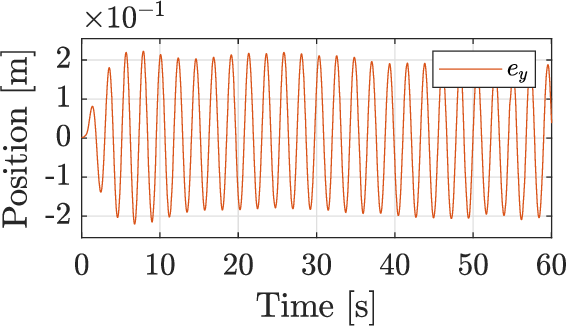}
    \end{subfigure}
    \hfill
    \begin{subfigure}[h]{0.48\columnwidth}
        \centering
        \includegraphics[width=\textwidth]{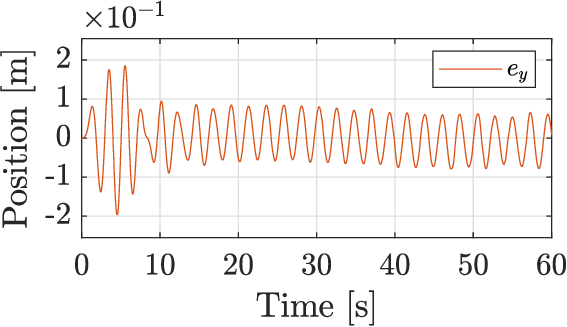}
    \end{subfigure}
    \caption{Experimental validation position error ${e_y}$ without and with learning compared}
    \label{fig:real_error_without_and_with_learning}
\end{figure}

\begin{figure}[h]
    \centering
    \includegraphics[width=\columnwidth]{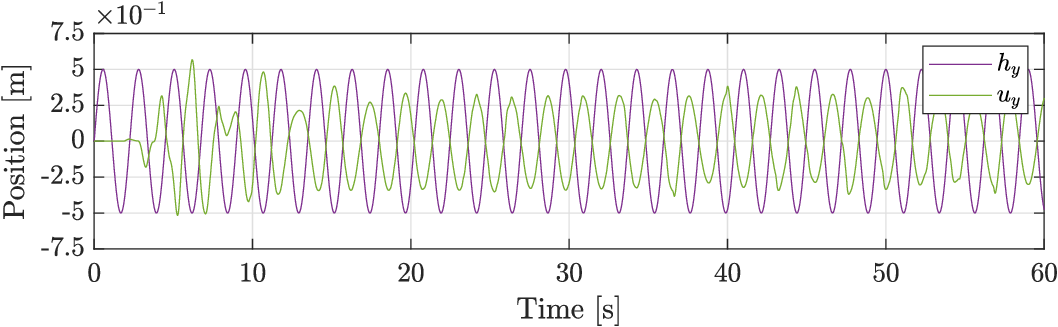}
    \caption{Experimental validation disturbance and adaptive input with learning enabled}
    \label{fig:real_disturb_vs_adaptive_in_with_learning}
\end{figure}

As seen from the position error ${e_y}$ shown in Figure~\ref{fig:real_error_without_and_with_learning}, the improvement is significant, as the nonparametric adaptive controller learns and cancels the disturbance. The improvement is quantified in Table \ref{tab:real_tracking_error_metrics}, and the learned control input from the nonparametric adaptive controller compared to the disturbance in the $y$-direction is shown in Figure~\ref{fig:real_disturb_vs_adaptive_in_with_learning}.

\begin{table}[h]
\vspace{3mm}
\caption{Tracking error metrics - Experimental validation}
\centering
\label{tab:real_tracking_error_metrics}
\begin{tabular}{@{\extracolsep\fill}m{38pt}ccc}
\toprule
Metric & W/o learn. & With learn. & Improvement [\%]\\
\midrule
MSE & $1.93 \cdot 10^{-2}$ & $3.66 \cdot 10^{-3}$ & $81.05$\\
MAE & $1.24 \cdot 10^{-1}$ & $5.16 \cdot 10^{-2}$ & $58.47$\\
\bottomrule
\end{tabular}
\end{table}

\section{Conclusion}\label{sec:conclusion}

A novel control algorithm has been presented for the automatic control of an offshore crane. The control algorithm uses a novel Cartesian model of a crane to design a tracking controller based on partial feedback linearization. The controller stabilizes the crane payload and tracks the reference trajectory, eliminating the need for a cascade of separate stabilizing and tracking controllers. Formal proofs have been presented which show that the proposed controller achieves uniformly ultimately bounded tracking errors. The Cartesian formulation allows the use of the novel nonparametric adaptive controller for disturbance rejection, such as wave disturbances, making the approach particularly relevant for enhancing the safety and efficiency of offshore crane operations.

Simulations showed that the controller is more accurate for trajectory tracking than an angular formulation. The tracking performance, as measured by the MSE of the tracking error, is improved by $99.34\%$ with a comparable velocity and acceleration of the suspension point. The nonparametric adaptive controller has been tested in simulation and experiments on an industrial robot. The tracking error MSE improved by $86.83\%$ in the simulation and $81.05\%$ in the experiments. This shows that the proposed controller significantly improves tracking performance when subject to disturbances.


\begin{ack}                               
The authors would like to acknowledge and thank Master Student Thomas Storvik for his contribution to producing the experimental lab setup and software.
\end{ack}


\bibliographystyle{plain}
\bibliography{utilities/refs}


\appendix

\section{Feature map for the Gaussian kernel}\label{sec:appendix_gaussian_kernel}

A feature map for the Gaussian kernel
\begin{equation}
    k(\x,\z) = \exp\left(-\frac{(\x-\z)^\tr(\x-\z)}{2\sigma^2}\right)
\end{equation}
is derived in \cite{Shashua2009} from
\begin{align}
    k(\x,\z) &= \exp\left(-\frac{\x^\tr\x}{2\sigma^2}\right) \exp\left(-\frac{\z^\tr\z}{2\sigma^2}\right)\exp\left(\frac{\x^\tr\z}{\sigma^2}\right)\nonumber \\
            &= \exp\left(-\frac{\x^\tr\x}{2\sigma^2}\right)\exp\left(-\frac{\z^\tr\z}{2\sigma^2}\right)\sum_{k=0}^\infty \frac{(\x^\tr\z)^k}{k!}
\end{align}
The term $(\x^\tr\z)^k$ gives
\begin{align}
    (\x^\tr\z)^k &= (x_1z_1 + \ldots + x_nz_n)^k \nonumber \\
                &= \sum_{k_1+\ldots + k_n = k}\frac{k!}{k_1!\ldots k_n!}(x_1z_1)^{k_1}\ldots (x_nz_n)^{k_n}\nonumber \\
                &= \sum_{k_1+\ldots + k_n = k}\sqrt{\frac{k!}{k_1!\ldots k_n!}} x_1^{k_1}\ldots x_n^{k_n} \sqrt{\frac{k!}{k_1!\ldots k_n!}}z_1^{k_1}\ldots z_n^{k_n}
\end{align}
where the second equality is due to the binomial theorem. The kernel can therefore be written as
\begin{align}
    k(\x,\z) &= \sum_{k=0}^\infty\left(\sum_{k_1+\ldots + k_n = k}\frac{\exp\left(-\frac{\x^\tr\x}{2\sigma^2}\right)}{\sqrt{k_1!\ldots k_n!}}x_1^{k_1}\ldots x_n^{k_n}\frac{\exp\left(-\frac{\z^\tr\z}{2\sigma^2}\right)}{\sqrt{k_1!\ldots k_n!}}z_1^{k_1}\ldots z_n^{k_n}\right)
\end{align}
Define the infinite-dimensional feature map $\phiv = [\phi_0,\phi_1,\phi_2\ldots]^\tr$ by the components
\begin{equation}
    \phi_k(\x) = \frac{\exp\left(-\frac{\x^\tr\x}{2\sigma^2}\right)}{\sqrt{k_1!\ldots k_n!}}x_1^{k_1}\ldots x_n^{k_n}, \quad k = 0,1,2\ldots
\end{equation}
where $k_1+\ldots + k_n = k$ and $k_1,\ldots k_n\geq 0$. This is a feature map for the Gaussian kernel since
\begin{equation}
    k(\x,\z) = \sum_{k=0}^\infty \phi_k(\x)^\tr\phi_k(\z) = \phiv(\x)^\tr\phiv(\z)
\end{equation}

\section{Crane model in Cartesian coordinates}\label{sec:appendix_cartesian_modeling}

\subsection{Kane's equations of motion for a spherical pendulum}

The Cartesian model is derived from the dynamic model using angular coordinates \cite{Tysse2022} by introducing a change of coordinates. The inertial frame $n$ is defined with the $z$-axis pointing upwards. The body-fixed frame $b$ is defined with the $z$ axis along the crane wire. The rotation from frame $n$ to frame $b$ is given by the Euler angles $\phi_x$ about the $x$ axis of the $n$ frame followed by a rotation $\phi_y$ about the resulting $y$ axis. The rotation matrix is then $\boldR^n_b = \boldR_x(\phi_x)\boldR_y(\phi_y)$. This gives
\beq
\boldR^n_b = \ba{ccc} 1 & 0 & 0 \\ 0 & c_x & -s_x \\ 0 & s_x & c_x\ea
\ba{ccc} c_y & 0 & s_y \\ 0 & 1 & 0 \\ -s_y & 0 & c_y\ea
= \ba{ccc} c_y & 0 & s_y \\ s_xs_y & c_x & -s_xc_y \\ -c_xs_y & s_x & c_xc_y\ea
\eeq

The position of the crane tip in the coordinates of $n$ is $\rv_0^n$ and the position of the mass is 
\beq
\rv^n = \rv_0^n + \boldR^n_b \rv_r^b
\eeq
where $\rv_r^b = [0,0,-L]^\tr$.
The velocity is $\vv^n = \dot\rv^n$ and the acceleration is $\av^n = \ddot\rv^n$. 
The coordinate expressions are 
\beq
\rv^n = \ba{c} x \\ y \\ z\ea
= \ba{c} x_0 - s_y L  \\ y_0 + s_x c_yL \\ z_0 - c_xc_yL \ea
\eeq 
and 
\beq\label{doublePendulumVelocity_components_angles}
\vv^n =
\ba{c} \dot x_0 - c_y\dot\phi_y L  - s_y \dot L\\ 
\dot y_0 + c_xc_y\dot\phi_xL - s_xs_y\dot\phi_yL + s_xc_y\dot L\\ 
\dot z_0 + s_xc_y\dot\phi_xL + c_xs_y\dot\phi_yL - c_xc_y\dot L\ea
\eeq
The acceleration is then found by differentiation of the velocity components to be
\begin{align}
\ddot x &= \ddot x_0 - c_y\ddot\phi_y L + s_y\dot\phi_y^2L 
- 2c_y\dot\phi_y\dot L - s_y \ddot L  
\\ 
\ddot y &= \ddot y_0 - Ls_xc_y(\dot\phi_x^2+\dot\phi_y^2) - 2Lc_xs_y\dot\phi_x\dot\phi_y
+ Lc_xc_y\ddot\phi_x - Ls_xs_y\ddot\phi_y \\
&\quad \quad - 2\dot L(-c_xc_y\dot\phi_x + s_xs_y\dot\phi_y) + \ddot L s_xc_y
\\ 
\ddot z &= \ddot z_0 + Lc_xc_y(\dot\phi_x^2+\dot\phi_y^2) - 2Ls_xs_y\dot\phi_x\dot\phi_y
+ Ls_xc_y\ddot\phi_x + Lc_xs_y\ddot\phi_y \\
&\quad \quad + 2\dot L(s_xc_y\dot\phi_x + c_xs_y\dot\phi_y) - \ddot L c_xc_y
\end{align}

The partial velocities with respect to the generalized speeds $(\dot\phi_x,\dot\phi_y)$, which are used in the development of Kane's equation of motion, are found from \eqref{doublePendulumVelocity_components_angles} to be
\begin{align}
\vv_1 &= \frac{\partial \vv^n}{\partial \dot\phi_x} 
= \ba{c} 0 \\ Lc_xc_y \\ Ls_xc_y \ea \\
\vv_2 &= \frac{\partial \vv^n}{\partial \dot\phi_y} 
= \ba{c} -Lc_y \\ -Ls_xs_y \\ Lc_xs_y \ea
\end{align}
Kane's equations of motion are then found from
\begin{align}
\vv_1^\tr(-m\av^n + m\g^n + \F) &= 0
\label{eq:Kane_sperical_pendulum_inner_product_1}\\
\vv_2^\tr(-m\av^n + m\g^n + \F) &= 0
\label{eq:Kane_sperical_pendulum_inner_product_2}
\end{align}
where $\F = [F_x,F_y,F_z]^\tr$ is the external force acting on the load and $\g = [0,0,-g]^\tr$ is the acceleration of gravity, where $g = \SI{9.81}{\metre\per\second\squared}$. After some simplifications, this gives
\begin{align}
mL c_y\left(-c_x \ddot y_0 - s_x \ddot z_0 + 2\dot Lc_y\dot\phi_x + 2L s_y\dot\phi_x\dot\phi_y - L c_y\ddot\phi_x  \right) \quad &
\nonumber \\
- L  s_xc_y mg L + Lc_xc_yF_y + Ls_xc_yF_z &= 0 \\
mL \left(c_y \ddot x_0 + \ddot y_0 s_xs_y - \ddot z_0 c_xs_y - L s_yc_y \dot\phi_x^2 
- L\ddot\phi_y  \right) \quad &
\nonumber \\
- L  c_x s_ymg - Lc_yF_x - Ls_xs_yF_y + Lc_xs_yF_z &= 0
\end{align}
Division of the first equation by $mL^2c_y$ and the second by $mL^2$ gives
\begin{align}
\ddot\phi_xc_y + \omega_0^2s_x &= \frac{1}{L}\left(-\ddot y_0c_x - \ddot z_0s_x + 2\dot Lc_y\dot\phi_x\right)
+ 2s_y\dot\phi_x\dot\phi_y 
+ \frac{c_x}{mL}F_y + \frac{x_x}{mL}F_z
\label{Kane_spherical_ddphix}\\
\ddot\phi_y + \omega_0^2c_xs_y 
&= \frac{1}{L}\left(\ddot x_0c_y + \ddot y_0s_xs_y - \ddot z_0c_xs_y - 2\dot L\dot\phi_y\right)
- s_yc_y\dot\phi_x^2 
\nonumber \\
&\quad 
- \frac{c_y}{mL}F_x - \frac{s_xs_y}{mL}F_y 
+ \frac{c_xs_y}{mL}F_z
\label{Kane_spherical_ddphiy}
\end{align}

\subsection{Change of coordinates to Cartesian model}

Let 
\beq
\rv_r^n = \rv^n + \rv_0^n
\eeq
be the relative position of the mass with respect to the crane tip. This is written in coordinate form as 
\beq
\ba{c} x_r \\ y_r \\ z_r\ea
= \ba{c} x-x_0  \\ y-y_0 \\ z-z_0 \ea
\eeq 
The vertical component of the cable length is 
\begin{equation}
    L_z = - z_r = \sqrt{L^2 - x_r^2 - y_r^2}
\end{equation}
where it is assumed that $z_r < 0$.

The relative velocity is $\vv_r^n = \dot\rv_r^n$ and the relative acceleration is $\av_r^n = \ddot\rv_r^n$. Then   
\beq\label{doublePendulumPosition_components}
\rv_r^n = \ba{c} x_r \\ y_r \\ z_r\ea
= \ba{c} - s_y L  \\ s_x c_yL \\ - c_xc_yL \ea
\eeq 
and
\beq\label{doublePendulumVelocity_components}
\vv_r^n = \ba{c} \dot x_r \\ \dot y_r \\ \dot z_r\ea =
\ba{c} - c_y\dot\phi_y L  - s_y \dot L\\ 
c_xc_y\dot\phi_xL - s_xs_y\dot\phi_yL + s_xc_y\dot L\\ 
s_xc_y\dot\phi_xL + c_xs_y\dot\phi_yL - c_xc_y\dot L\ea
\eeq
In the following it is assumed that $\dot L = 0$, so that $x_r\dot x_r + y_r\dot y_r + z_r\dot z_r = 0$. Then 
\beq
\ba{c} \dot x_r \\ \dot y_r  \ea
=  
\underbrace{
\ba{cc} 0 & -c_y L 
\\ c_x c_yL & -s_xs_yL 
 \ea
}_{\A}
\ba{c} \dot\phi_x  \\ \dot\phi_y  \ea
\eeq
and
\beq
\ba{c} \dot\phi_x  \\ \dot\phi_y  \ea
=  
\underbrace{
\ba{cc} -\frac{s_xs_y}{c_xc_y^2L} & \frac{1}{c_xc_yL}
\\ -\frac{1}{c_yL} & 0 
 \ea
}_{\inv{\A}}
\ba{c} \dot x_r \\ \dot y_r  \ea
\eeq

From the position coordinate expressions \eqref{doublePendulumPosition_components}, it is seen that  
\begin {align}
s_y &= -\frac{x_r}{L} 
\label{Cartesian-trigonometric-expression-1}\\
s_xc_y &= \frac{y_r}{L} \\
c_xc_y &= -\frac{z_r}{L}
\end{align}
It is noted that 
\beq
L^2 = x_r^2 + y_r^2 + z_r^2
\eeq
This gives 
\begin{align}
c_y &= \sqrt{1-s_y^2} = \sqrt{1 - \frac{x_r^2}{L^2}} = \frac{\sqrt{L^2 - x_r^2}}{L}
= \frac{\sqrt{y_r^2 + z_r^2}}{L}, \quad \phi_y < \frac{\pi}{2}
\\
s_xs_y &= s_xc_y\frac{1}{c_y}s_y 
= \frac{y_r}{L}\frac{L}{\sqrt{y_r^2 + z_r^2}} \frac{-x_r}{L}
= -\frac{x_ry_r}{L\sqrt{y_r^2 + z_r^2}}
\\
c_xs_y &= c_xc_y\frac{1}{c_y}s_y 
= \frac{z_r}{\sqrt{y_r^2 + z_r^2}}\frac{x_r}{L}
\\
s_xc_x &= \frac{(s_xc_y)(c_xc_y)}{c_y^2} = -\frac{y_rz_r}{y_r^2 + z_r^2}\\
\frac{s_xs_y}{c_xc_y^2} &= s_xs_y \frac{1}{c_xc_y}\frac{1}{c_y}
= -\frac{x_ry_r}{L\sqrt{y_r^2 + z_r^2}} \frac{-L}{z_r}\frac{L}{\sqrt{y_r^2 + z_r^2}}
= \frac{x_ry_rL}{z_r(y_r^2 + z_r^2)}
\label{Cartesian-trigonometric-expression-6}
\end{align}
This gives
\beq
\A = \ba{cc} 0 & -c_y L \\ c_x c_yL & -s_xs_yL 
 \ea
= \ba{cc} 0 & -\sqrt{y_r^2 + z_r^2} \\ 
-z_r & \frac{x_ry_r}{\sqrt{y_r^2 + z_r^2}} 
 \ea
\eeq

The determinant of the Jacobian $\A$ is $\det(\A) = 1/(c_xc_y^2L^2)$
which means that $\A$ is nonsingular whenever $c_x\neq 0$ and $c_y \neq 0$. The inverse matrix is
\begin{align}
\inv{\A} &= \ba{cc} -\frac{s_xs_y}{c_xc_y^2L} & \frac{1}{c_xc_yL}
\\ -\frac{1}{c_yL} & 0 \ea 
= \ba{cc} 
- \frac{x_ry_r}{z_r(y_r^2+z_r^2)} & -\frac{1}{z_r} \\ 
- \frac{1}{\sqrt{y_r^2 + z_r^2}} & 0 
\ea
\end{align}
which is verified by direct calculation. 

It follows that 
\begin{align}
\dot\phi_x^2 &= \left(\frac{x_ry_r}{z_r(y_r^2+z_r^2)} \dot x_r + \frac{1}{z_r} \dot y_r  \right)^2 \\
&= \frac{x_r^2y_r^2\dot x_r^2 }{z_r^2(y_r^2+z_r^2)^2}
+ 2\frac{x_ry_r\dot x_r\dot y_r}{z_r^2(y_r^2+z_r^2)}
+ \frac{\dot y_r^2}{z_r^2} \\
\dot\phi_y^2 &= \frac{\dot x_r^2}{y_r^2 + z_r^2}
\end{align}

The accelerations are given by 
\begin{align}
\ddot x &=  \ddot x_0 - Lc_y\ddot\phi_y  + s_y\dot\phi_y^2L   
\\ 
\ddot y &= \ddot y_0 + Lc_xc_y\ddot\phi_x - Ls_xs_y\ddot\phi_y  - Ls_xc_y(\dot\phi_x^2+\dot\phi_y^2) - 2Lc_xs_y\dot\phi_x\dot\phi_y 
\end{align}

The equations of motion in terms of the Euler angles are given by
\begin{align}
c_y\ddot\phi_x  &= \frac{1}{L}\left(- L\omega_0^2s_x - \ddot y_0c_x + 2Ls_y\dot\phi_x\dot\phi_y 
+ \frac{c_x}{m}F_y + \frac{s_x}{m}F_z
\right)
\\
\ddot\phi_y 
&= \frac{1}{L}\left(- L\omega_0^2c_xs_y +\ddot x_0c_y + \ddot y_0s_xs_y 
- Ls_yc_y\dot\phi_x^2 -\frac{c_y}{m}F_x - \frac{s_xs_y}{m}F_y - \frac{c_xs_y}{m}F_z\right)
\end{align}
Insertion of the equations of motion in the expressions for the accelerations and simplification using \eqref{Cartesian-trigonometric-expression-1}--\eqref{Cartesian-trigonometric-expression-6} gives for the $x$ direction
\begin{align}
\ddot x   &= \ddot x_0 -c_y\left(- L\omega_0^2c_xs_y +\ddot x_0c_y + \ddot y_0s_xs_y 
- Ls_yc_y\dot\phi_x^2 -\frac{c_y}{m}F_x - \frac{s_xs_y}{m}F_y + \frac{c_xs_y}{m}F_z \right) 
\nonumber\\
&\quad \quad + s_y\dot\phi_y^2L  \nonumber\\
&= \omega_0^2L c_xc_ys_y + s_y^2\ddot x_0 - \ddot y_0s_xc_ys_y + Ls_yc_y^2\dot\phi_x^2 + s_y\dot\phi_y^2L 
\nonumber\\
&\quad \quad  +\frac{c_y^2}{m}F_x + \frac{s_xc_ys_y}{m}F_y + \frac{c_xc_ys_y}{m}F_z
\end{align}
For the $y$-direction, the equation of motion is 
\begin{align}
\ddot y &= \ddot y_0
+ c_x\left(- L\omega_0^2s_x - \ddot y_0c_x + 2Ls_y\dot\phi_x\dot\phi_y + \frac{c_x}{m}F_y + \frac{s_x}{m}F_z \right) 
\nonumber\\
&\quad \quad 
- s_xs_y\left(- L\omega_0^2c_xs_y +\ddot x_0c_y + \ddot y_0s_xs_y 
- Ls_yc_y\dot\phi_x^2 -\frac{c_y}{m}F_x - \frac{s_xs_y}{m}F_y - \frac{c_xs_y}{m}F_z \right) 
\nonumber\\
&\quad \quad - Ls_xc_y\left(\dot\phi_x^2+\dot\phi_y^2\right) - 2Lc_xs_y\dot\phi_x\dot\phi_y 
\\ 
&= - s_xc_x(1 - s_y^2)L\omega_0^2
\nonumber\\
&\quad \quad
- (s_xc_y)s_y\ddot x_0 + (1 - c_x^2 - s_x^2s_y^2)\ddot y_0  \nonumber\\ 
&\quad \quad 
- Ls_xc_y (1 - s_y^2) \dot\phi_x^2 - Ls_xc_y\dot\phi_y^2 
\nonumber\\
&\quad\quad
+ c_x\left(\frac{c_x}{m}F_y + \frac{s_x}{m}F_z\right) 
- s_xs_y\left(-\frac{c_y}{m}F_x - \frac{s_xs_y}{m}F_y - \frac{c_xs_y}{m}F_z\right)
\\
&= s_xc_xc_y^2L\omega_0^2
- (s_xc_y)s_y\ddot x_0 + (s_xc_y)^2\ddot y_0 - Ls_xc_y c_y^2 \dot\phi_x^2 - Ls_xc_y\dot\phi_y^2 
\nonumber\\
&\quad\quad
+\frac{s_xs_yc_y}{m}F_x + \frac{s_x^2s_y^2 + c_x^2}{m}F_y + \frac{s_xc_x(1+s_y^2)}{m}F_z
\end{align} 
The accelerations are then rewritten in the form 
\begin{align}
\ddot x &= (c_xc_y)s_yL\omega_0^2  + s_y^2\ddot x_0 - (s_xc_y)s_y\ddot y_0 + Ls_yc_y^2\dot\phi_x^2 + s_y\dot\phi_y^2L 
\nonumber\\
&\quad \quad  +\frac{c_y^2}{m}F_x + \frac{(s_xc_y)s_y}{m}F_y + \frac{(c_xs_y)c_y}{m}F_z
\\
\ddot y &= (s_xc_x)c_y^2L\omega_0^2
- (s_xc_y)s_y\ddot x_0 + (s_xc_y)^2\ddot y_0 - L(s_xc_y)c_y^2 \dot\phi_x^2 - L(s_xc_y)\dot\phi_y^2 
\nonumber\\
&\quad\quad
+\frac{(s_xc_y)s_y}{m}F_x + \frac{s_x^2s_y^2 + c_x^2}{m}F_y + \frac{s_xc_x(1+s_y^2)}{m}F_z
\end{align} 
to make it easy to use \eqref{Cartesian-trigonometric-expression-1}--\eqref{Cartesian-trigonometric-expression-6}. This leads to
\begin{align}
\ddot x + \frac{L_z}{L}\omega_0^2 x &= \frac{L_z}{L}\omega_0^2 x_0 
 + \frac{x_r^2}{L^2}\ddot x_0 + \frac{x_ry_r}{L^2}\ddot y_0 
 - \frac{x_r(y_r^2 + z_r^2)}{L^2}\dot\phi_x^2 - x_r\dot\phi_y^2
\nonumber\\
&\quad \quad +\frac{y_r^2+z_r^2}{mL^2}F_x - \frac{x_ry_r}{mL^2}F_y + \frac{x_rz_r}{L^2}F_z\\
\ddot y + \frac{L_z}{L}\omega_0^2 y &= \frac{L_z}{L}\omega_0^2 y_0
+\frac{x_ry_r}{L^2}\ddot x_0 + \frac{y_r^2}{L^2}\ddot y_0 
 - \frac{y_r(y_r^2+z_r^2)}{L^2}\dot\phi_x^2 - y_r\dot\phi_y^2
\nonumber\\ &\quad \quad 
-\frac{x_ry_r}{mL^2}F_x + \frac{x_r^2 + z_r^2}{mL^2}F_y - \frac{y_rz_r}{L^2}F_z
\end{align}
Insertion of 
\begin{align}
\dot\phi_x^2 &= \frac{x_r^2y_r^2\dot x_r^2 }{z_r^2(y_r^2+z_r^2)^2}
+ 2\frac{x_ry_r\dot x_r\dot y_r}{z_r^2(y_r^2+z_r^2)}
+ \frac{\dot y_r^2}{z_r^2} \\
\dot\phi_y^2 &= \frac{\dot x_r^2}{y_r^2 + z_r^2}
\end{align}
gives
\begin{align}
\ddot x + \frac{L_z}{L}\omega_0^2 x &= \frac{L_z}{L}\omega_0^2 x_0 
+ \frac{x_r^2}{L^2}\ddot x_0 + \frac{x_ry_r}{L^2}\ddot y_0 
- \frac{x_r\dot x_r^2}{y_r^2 + z_r^2}
\nonumber\\
&\quad \quad- \frac{x_r(y_r^2 + z_r^2)}{L^2}\left( \frac{x_r^2y_r^2\dot x_r^2 }{z_r^2(y_r^2+z_r^2)^2}
+ 2\frac{x_ry_r\dot x_r\dot y_r}{z_r^2(y_r^2+z_r^2)}
+ \frac{\dot y_r^2}{z_r^2}\right)
\\
\ddot y + \frac{L_z}{L}\omega_0^2 y &= \frac{L_z}{L}\omega_0^2 y_0
+\frac{x_ry_r}{L^2}\ddot x_0 + \frac{y_r^2}{L^2}\ddot y_0 
- \frac{ y_r \dot x_r^2}{y_r^2 + z_r^2}
\nonumber\\
&\quad \quad  - \frac{y_r(y_r^2+z_r^2)}{L^2}
\left( \frac{x_r^2y_r^2\dot x_r^2 }{z_r^2(y_r^2+z_r^2)^2}
+ 2\frac{x_ry_r\dot x_r\dot y_r}{z_r^2(y_r^2+z_r^2)}
+ \frac{\dot y_r^2}{z_r^2}\right) 
\end{align}
and, finally 
\begin{align}
\ddot x + \frac{L_z}{L}\omega_0^2 x &= \frac{L_z}{L}\omega_0^2 x_0 
 + \frac{x_r^2}{L^2}\ddot x_0 + \frac{x_ry_r}{L^2}\ddot y_0 - \frac{x_r^3y_r^2\dot x_r^2 }{L^2L_z^2(y_r^2+z_r^2)} 
\nonumber\\
&\quad \quad 
- 2\frac{x_r^2y_r\dot x_r\dot y_r}{L^2L_z^2}
-\frac{x_r\dot y_r^2 (y_r^2 + L_z^2)}{L^2L_z^2}
- \frac{x_r\dot x_r^2}{y_r^2 + z_r^2}
\\
\ddot y + \frac{L_z}{L}\omega_0^2 y &= \frac{L_z}{L}\omega_0^2 y_0
+\frac{x_ry_r}{L^2}\ddot x_0 + \frac{y_r^2}{L^2}\ddot y_0 - \frac{x_r^2y_r^3\dot x_r^2 }{L^2L_z^2(y_r^2+z_r^2)}
\nonumber\\
&\quad \quad 
- 2\frac{x_ry_r^2\dot x_r\dot y_r}{L^2L_z^2}
- \frac{y_r\dot y_r^2(y_r^2+z_r^2)}{L^2L_z^2}
- \frac{ y_r \dot x_r^2}{y_r^2 + z_r^2}
\end{align}


\end{document}